\documentclass[default,iicol]{sn-jnl}

\jyear{2024}

\theoremstyle{thmstyleone}

\theoremstyle{thmstyletwo}

\theoremstyle{thmstylethree}

\usepackage{overpic}
\usepackage{color}
\usepackage{colortbl}
\usepackage{marvosym}
\usepackage{enumerate}
\usepackage{wrapfig}
\usepackage{booktabs}
\usepackage{multicol,multirow}

\usepackage{url}
\usepackage{xcolor}
\usepackage{hyperref}

\newcommand{\myrotate}[1]{ \rotatebox{90}{#1}}

\newcommand{\secref}[1]{Sec.~\ref{#1}}

\def\ie{i.e.}

\def\etal{et al.}

\definecolor{mygray}{gray}{.9}
\definecolor{zhh}{RGB}{219, 68, 55}
\definecolor{myGreen}{RGB}{15, 157, 88}
\definecolor{myBlue}{RGB}{66, 133, 244}

\newcommand{\blu}[1]{{\textit{#1}}}
\newcommand{\rev}[1]{{\textbf{#1}}}
\newcommand{\gre}[1]{{\underline{#1}}}

\graphicspath{{./Imgs/}{../../figure/}}

\graphicspath{{Imgs/}}

\newcommand\blfootnote[1]{
    \begingroup
    \renewcommand\thefootnote{}\footnote{#1}
    \addtocounter{footnote}{-1}
    \endgroup
}

\begin{document}

\title[PolypSegSurvey]{\textbf{A Survey on Deep Learning for Polyp Segmentation: Techniques, Challenges and Future Trends}}

\author[1]{\fnm{Jiaxin} \sur{Mei}} 
\author[1]{\fnm{Tao} \sur{Zhou}$\textsuperscript{\Letter}$} 
\author[1]{\fnm{Kaiwen} \sur{Huang}} 
\author[1]{\fnm{Yizhe} \sur{Zhang}}
\author[2]{\fnm{Yi} \sur{Zhou}} 
\author[1]{\fnm{Ye} \sur{Wu}} 
\author[3]{\fnm{Huazhu} \sur{Fu}}

\affil[1]{\orgdiv{Key Laboratory of Intelligent Perception and Systems for High-Dimensional Information of Ministry of Education, and School of Computer Science and Engineering, Nanjing University of Science and Technology}, \orgaddress{\city{Nanjing}, \country{China}}}

\affil[2]{\orgdiv{School of Computer Science and Engineering}, \orgname{Southeast University}, \orgaddress{\city{Nanjing}, \country{China}}}

\affil[3]{\orgdiv{Institute of High Performance Computing, A*STAR, \country{Singapore}}}


\abstract{
Early detection and assessment of polyps play a crucial role in the prevention and treatment of colorectal cancer (CRC). Polyp segmentation provides an effective solution to assist clinicians in accurately locating and segmenting polyp regions. In the past, people often relied on manually extracted lower-level features such as color, texture, and shape, which often had problems capturing global context and lacked robustness to complex scenarios. With the advent of deep learning, more and more medical image segmentation algorithms based on deep learning networks have emerged, making significant progress in the field. This paper provides a comprehensive review of polyp segmentation algorithms. We first review some traditional algorithms based on manually extracted features and deep segmentation algorithms, and then describe benchmark datasets related to the topic. Specifically, we carry out a comprehensive evaluation of recent deep learning models and results based on polyp size, taking into account the focus of research topics and differences in network structures. Finally, we discuss the challenges of polyp segmentation and future trends in the field. 
}


\maketitle

\section{Introduction}
\blfootnote{$\textsuperscript{\Letter}$ Correspondence to: taozhou.ai@gmail.com.}

Polyp segmentation aims to automatically identify and segment polyp regions within the colon. Its primary objective is to assist clinical doctors in efficiently and accurately locating and delineating these regions, providing vital support for early diagnosis and treatment of colorectal cancer (CRC)~\cite{guo2020learn}. Polyps exhibit varying sizes and shapes at different stages of development~\cite{yang2020colon,zhou2024uncertainty}, and their precise segmentation poses challenges due to their strong adherence to adjacent organs or mucosa~\cite{fan2021concealed}. 

In the early stage, polyp segmentation primarily relied on manually extracted features~\cite{tajbakhsh2015automated,iwahori2017automatic}. For example, Tajbakhsh \etal~\cite{tajbakhsh2015automated} proposed a method that utilized shape features and surrounding environmental information to automatically detect polyps in colonoscopy videos. Iwahori \etal~\cite{iwahori2017automatic} utilized edge and color information to generate a likelihood map and extracted the directional gradient histogram features. However, relying on manually extracted low-level features for segmentation tasks makes it difficult to handle complex scenarios and does not effectively utilize global contextual information~\cite{bernal2015wm}. 

\begin{figure*}[t]
\centering
\begin{overpic}[width=1.0\linewidth]{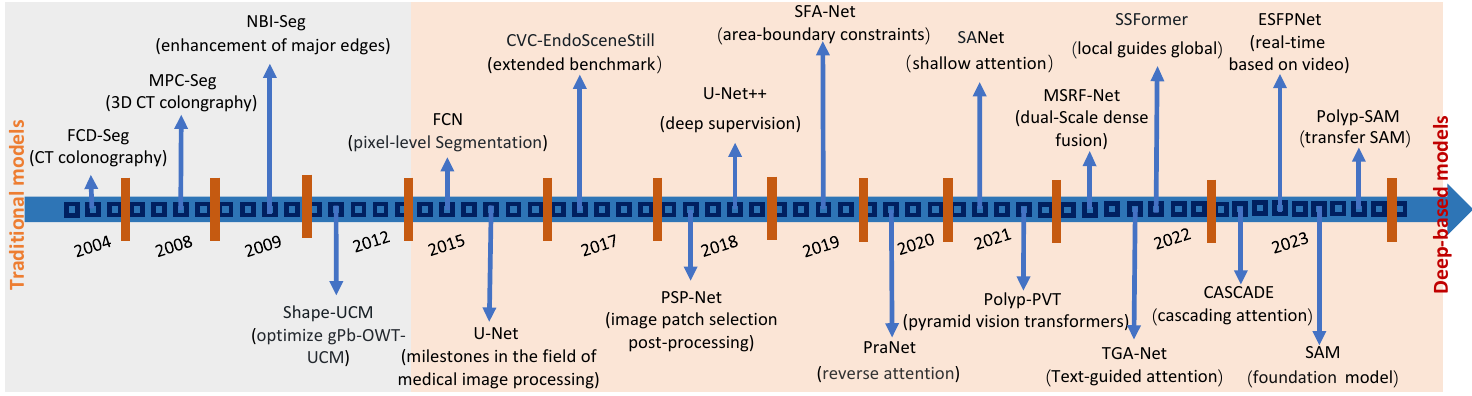}

\put(8.3,17.3){\footnotesize ~\cite{yao2004colonic}}  
\put(14.3,21.1){\footnotesize ~\cite{lu2008accurate}} 
\put(20.3,24.9){\footnotesize ~\cite{gross2009polyp}} 
\put(24.9,5.3){\footnotesize ~\cite{ganz2012automatic}}
\put(31.0,18.4){\footnotesize ~\cite{long2015fully}}
\put(34.7,3.9){\footnotesize ~\cite{ronneberger2015u}}
\put(44.0,23.9){\footnotesize ~\cite{vazquez2017benchmark}}
\put(48.6,6.4){\footnotesize ~\cite{akbari2018polyp}}
\put(51.6,20.3){\footnotesize ~\cite{zhou2018unet++}}
\put(57.6,25.6){\footnotesize ~\cite{fang2019selective}}
\put(61.8,3.0){\footnotesize ~\cite{fan2020pranet}}
\put(67.8,24.1){\footnotesize ~\cite{wei2021shallow}}
\put(71.6,6.2){\footnotesize ~\cite{dong2021polyp}}
\put(75.3,20.0){\footnotesize ~\cite{srivastava2021msrf}}
\put(78.8,2.4){\footnotesize ~\cite{tomar2022tganet}}
\put(80.1,25.1){\footnotesize ~\cite{wang2022stepwise}}
\put(86.0,6.6){\footnotesize ~\cite{rahman2023medical}}
\put(89.2,25.4){\footnotesize ~\cite{chang2023esfpnet}}
\put(90.6,2.6){\footnotesize ~\cite{kirillov2023segment}}
\put(94.5,18.6){\footnotesize ~\cite{li2023polyp}}

\end{overpic}\vspace{-0.05cm}
\caption{ A brief chronology of polyp segmentation. Before 2015, methods relied on hand-crafted features combined with machine learning algorithms. Since 2015, U-Net~\cite{ronneberger2015u} and FCN~\cite{long2015fully} have significantly advanced the development of deep learning techniques in polyp segmentation. More details can be found in \secref{sec:models}.}\label{fig_timeline}\vspace{-0.25cm}

\end{figure*}

In recent years, deep learning-based polyp segmentation models have demonstrated impressive capabilities in locating and segmenting polyp regions. Based on the encoder-decoder architecture, some works focus on enhancing the output features of the encoder~\cite{wei2021shallow,zhao2021automatic}. In addition, the variation between datasets collected from different devices is also an issue~\cite{yang2021mutual}. After the emergence of the segment anything model (SAM)~\cite{kirillov2023segment}, fine-grained models based on universal segmentation networks have shown promising momentum in polyp segmentation~\cite{li2023polyp}. To provide a clearer depiction of the progress in polyp segmentation tasks, we show a concise timeline in Fig.~\ref{fig_timeline}. 

Existing surveys typically provide systematic statistics on polyp segmentation, localization, and detection methods~\cite{sanchez2020deep}, or review medical image segmentation across different anatomical regions, such as the abdomen, heart, brain, and lungs~\cite{XIAO2023104791,gupta2023review,qureshi2023medical,chowdhary2020segmentation,thisanke2023semantic,bennai2023multi}. In contrast to previous reviews on polyp segmentation or medical image segmentation, this paper aims to provide a comprehensive review of polyp segmentation methods. It covers both traditional algorithms and deep learning model-based approaches for polyp segmentation. 

We discuss traditional methods and deep learning-based methods separately. Based on the unique design of model structures, we categorize deep models into boundary-aware models, attention-aware models, and feature fusion models. We discuss the advantages associated with each method and provide valuable insights into their respective applicability and performance characteristics. In \secref{sec:models}, we review the existing polyp segmentation models from different aspects. In \secref{sec:dataset}, we summarize and provide detailed information and usage of the current publicly available datasets used for polyp segmentation. Then, we conduct a comprehensive assessment of the segmentation performance for polyp sizes and an analysis of the advantages and disadvantages of several representative polyp segmentation models in \secref{sec:evaluation}. After that, in \secref{sec:challenge}, we discuss the challenges and future trends for development in the field. Finally, we conclude the paper in \secref{sec:conclusion}.

\renewcommand\arraystretch{1.25}
\setlength{\tabcolsep}{0.5pt}
\begin{table*}
    \centering
    
    \caption{Summary of polyp segmentation methods (published from 2019 to 2021).}
    \tiny
    \scriptsize
    \label{tab:001}
    \centering
    \begin{tabular}{|p{0.4cm}|p{0.8cm}|p{2.5cm}|p{1.5cm}|p{1.9cm}|p{8.0cm}|p{0.8cm}|}
    
        \hline
         \centering\textbf{\#} & \centering\textbf{Year} & \centering\textbf{Method} & \centering\textbf{Pub.} & \centering\textbf{Backbone} & \centering\textbf{Description}  & \textbf{Code} \\ \hline
    1 & 2019 & SFA	~\cite{fang2019selective} & MICCAI & light UNet & Boundary-sensitive loss; selective feature aggregation & \href{link}{https://github.com/Yuqi-cuhk/Polyp-Seg} \\ \hline

    2 & 2019 & ResUNet++~\cite{jha2019resunet++} & ISM & ResUNet & Squeeze and excitation blocks; atrous spatial pyramid pooling(ASPP); attention blocks & \href{link}{https://github.com/DebeshJha/ResUNetPlusPlus} \\ \hline

    3 & 2020 & PolypSeg~\cite{zhong2020polypseg} & MICCAI & UNet & Improved attention mechanism; separable convolution & N/A \\ \hline

    4 & 2020 & ThresholdNet~\cite{guo2020learn} & TMI & DeepLabv3+ & Confidence-guided manifold mixup; threshold loss & \href{link}{https://github.com/Guo-Xiaoqing/ThresholdNet} \\ \hline

    5 & 2020 & ACSNet~\cite{zhang2020adaptive} & MICCAI & ResNet34 & Adaptively select; aggregate context features through channel attention & \href{link}{https://github.com/ReaFly/ACSNet} \\ \hline

    6 & 2020 & PraNet~\cite{fan2020pranet} & MICCAI & Res2Net & Parallel partial decoders; reverse attention & \href{link} {https://github.com/DengPingFan/PraNet}\\ \hline

    7 & 2021 & DDANet~\cite{tomar2021ddanet} & PR & ResUNet & Dual-decoder attention; out-of-training-set testing & \href{link}{https://github.com/nikhilroxtomar/DDANet} \\ \hline

    8 & 2021 & MSRF-Net~\cite{srivastava2021msrf} & JBHI  & N/A & Multi-scale residual fusion; dual-scale dense fusion & \href{link}{https://github.com/NoviceMAn-prog/MSRF-Net} \\ \hline

    9 & 2021 & HarDNet-MSEG~\cite{huang2021hardnet} & arxiv & HarDNet68 & Cascaded partial decoder; dense aggregation & \href{link}{https://github.com/james128333/HarDNet-MSEG} \\ \hline

    10 & 2021 & EU-Net~\cite{patel2021enhanced} & CRV &  ResNet34 & Semantic feature; adaptive global context module & \href{link}{https://github.com/rucv/Enhanced-U-Net} \\ \hline

    11 & 2021 & FANet~\cite{tomar2022fanet} & TNNLS & N/A & Feedback attention learning; iterative refining; embedded run-length encoding strategy & \href{link}{https://github.com/nikhilroxtomar/fanet}\\ \hline

    12 & 2021 & Polyp-PVT~\cite{dong2021polyp} & CAAI AIR & PVT & Cascaded fusion; camouflage identification; similarity aggregation & \href{link}{https://github.com/DengPingFan/Polyp-PVT} \\ \hline

    13 & 2021 & UACANet	~\cite{kim2021uacanet} & ACM MM & Res2Net & Parallel axial attention; uncertainty augmented context attention & \href{link}{https://github.com/plemeri/UACANet} \\ \hline

    14 & 2021 & C2FNet~\cite{sun2021context} & IJCAI & Res2Net-50 & Context-aware cross-level fusion; dual-branch global context & \href{link}{https://github.com/thograce/C2FNet} \\ \hline

    15 & 2021 & ResUNet++ + TTA + CRF~\cite{jha2021comprehensive} & JBHI & ResUNet & conditional random fields; test-time augmentation & \href{link}{https://github.com/DebeshJha/ResUNet-with-CRF-and-TTA} \\ \hline

    16 & 2021 & MPA-DA~\cite{yang2021mutual} & JBHI & ResNet-101 & Inter-prototype adaptation network; progressive self-training; disentangled reconstruction & \href{link}{https://github.com/CityU-AIM-Group/MPA-DA} \\ \hline

    17 & 2021 & TransFuse~\cite{zhang2021transfuse} & MICCAI &  ResNet-34 + DeiT-S  & self-attention; bilinear hadamard product; gated skip-connection & \href{link}{https://github.com/Rayicer/TransFuse} \\ \hline 

    18 & 2021 & SANet~\cite{wei2021shallow} & MICCAI & Res2Net & Color exchange; shallow attention & \href{link}{https://github.com/weijun88/sanet} \\ \hline

    19 & 2021 & STFT~\cite{wu2021multi} & MICCAI & ResNet-50 &  Spatial-temporal feature transformation; deformable convolutions and channel-aware attention & \href{link}{https://github.com/lingyunwu14/STFT} \\ \hline

    20 & 2021 & LOD-Net~\cite{cheng2021learnable} & MICCAI & ResNet + FPN &  Model the probability of each pixel being in the boundary region;  adaptive thresholding policy & \href{link}{https://github.com/midsdsy/LOD-Net} \\ \hline
    
    21 & 2021 & MSNet~\cite{zhao2021automatic} & MICCAI & Res2Net-50 &  Multi-scale subtraction; training-free network & \href{link}{https://github.com/Xiaoqi-Zhao-DLUT/MSNet} \\ \hline

    22 & 2021 & CCBANet~\cite{nguyen2021ccbanet} & MICCAI & ResNet34 &  Cascading context; balancing attention & \href{link}{https://github.com/ntcongvn/CCBANet} \\ \hline

    23 & 2021 & HRENet~\cite{shen2021hrenet} & MICCAI & ResNet34 & Hard region enhancement; adaptive feature aggregation; edge and structure consistency aware loss & \href{link}{https://github.com/CathySH/HRENet} \\ \hline
    
    24 & 2021 & GMSRF-Net~\cite{srivastava2022gmsrf} & ICPR & ResNet50 & Cross-multi-scale attention; multi-scale feature selection & \href{link}{https://github.com/NoviceMAn-prog/GMSRFNet} \\ \hline 
    
    \hline
    \end{tabular}
\end{table*}

\renewcommand\arraystretch{1.25}
\setlength{\tabcolsep}{0.5pt}
\begin{table*}
    \centering
    
    \caption{Summary of polyp segmentation methods (published from 2022 to 2023).}
    \scriptsize
    \label{tab:002}
    \begin{tabular}{|p{0.4cm}|p{0.8cm}|p{2.5cm}|p{1.5cm}|p{1.9cm}|p{8.0cm}|p{0.8cm}|}
    
        \hline
         \centering\textbf{\#} & \centering\textbf{Year} & \centering\textbf{Method} & \centering\textbf{Pub.} & \centering\textbf{Backbone} & \centering\textbf{Description}  & \textbf{Code} \\ \hline

    25 & 2022 & TGANet~\cite{tomar2022tganet} & MICCAI  & ResNet50 & Text-guided attention; weights the text-based embeddings & \href{link}{https://github.com/nikhilroxtomar/tganet} \\ \hline

    26 & 2022 & CaraNet	~\cite{lou2022caranet} & JMI & Res2Net & Context axial reverse attention & \href{link}{https://github.com/AngeLouCN/CaraNet} \\ \hline 

    27 & 2022 & MSRAformer~\cite{wu2022msraformer} & CBM &  Swin Transformer & multi-scale spatial reverse attention & \href{link}{https://github.com/ChengLong1222/MSRAformer-main} \\ \hline

    28 & 2022 & HSNet~\cite{zhang2022hsnet} & CBM & PVTv2 & Cross-semantic attention; hybrid semantic complementary; multi-scale prediction & \href{link}{https://github.com/baiboat/HSNet} \\ \hline

    29 & 2022 & FuzzyNet~\cite{patel2022fuzzynet} & NeurIPS & Res2Net/ ConvNext/ PVT &  fuzzy attention; focus on the blurry pixels & \href{link}{https://github.com/krushi1992/FuzzyNet} \\ \hline
    
    30 & 2022 & LDNet~\cite{zhang2022lesion} & MICCAI  & Res2Net & Lesion-aware dynamic kernel; self-attention & \href{link}{https://github.com/ReaFly/LDNet} \\ \hline
        
    31 & 2022 & HarDNet-DFUS~\cite{liao2022hardnet} & arxiv &  HarDNetV2 & Real-time model; enhanced backbone & \href{link}{https://github.com/YuWenLo/HarDNet-DFUS} \\ \hline

    32 & 2022 & BDG-Net~\cite{qiu2022bdg} & SPIE MI & EfficientNet-B5 & Boundary distribution guided; boundary distribution generate & \href{link}{https://github.com/zihuanqiu/BDG-Net} \\ \hline

    33 & 2022 & ColonFormer~\cite{duc2022colonformer} & Access &  MiT & Integrate a hierarchical Transformer and a hierarchical pyramid CNN; residual axial attention & \href{link}{https://github.com/ducnt9907/ColonFormer} \\ \hline

    34 & 2022 & FCBFormer~\cite{sanderson2022fcn} & MIUA  &  PVTv2 & Improved progressive locality decoder;  fully convolutional branch + Transformer branch & \href{link}{https://github.com/ESandML/FCBFormer} \\ \hline

    35 & 2022 & DCRNet~\cite{yin2022duplex} & ISBI  & ResNet-34 & Duplex contextual relation network; cross-image contextual relations & \href{link}{https://github.com/PRIS-CV/DCRNet} \\ \hline

    36 & 2022 & SSFormer~\cite{wang2022stepwise} & MICCAI  & PVTv2 & Aggregate local and global features stepwise & \href{link}{https://github.com/Qiming-Huang/ssformer} \\ \hline
    
    37 & 2022 & DuAT~\cite{tang2023duat} & PRCV & PVT &  Dual-aggregation transformer; global-to-local spatial aggregation; selective boundary aggregation & \href{link}{https://github.com/Barrett-python/DuAT} \\ \hline
    
    38 & 2022 & LAPFormer~\cite{nguyen2022lapformer} & arxiv  & MiT-B1 & Hierarchical transformer encoder and CNN decoder; progressive feature fusion & N/A \\ \hline

    39 & 2022 & PPFormer~\cite{cai2022using} &  MICCAI  & CvT &  Shallow CNN encoder; deep Transformer-based encoder & N/A \\ \hline

    40 & 2022 & BSCA-Net~\cite{lin2022bsca} & PR  & Res2Net &  Bit-plane slicing information; segmentation squeeze bottleneck union module; multi-path connection attention & N/A \\ \hline

    41 & 2022 & BoxPolyp~\cite{wei2022boxpolyp} &  MICCAI  & Res2Net/ PVT  & Box annotations; fusion filter sampling module & N/A \\ \hline

    42 & 2022 & ICBNet~\cite{xiao2022icbnet} & BIBM & PVT &  Iterative feedback learning strategy; context and boundary-aware information & N/A \\ \hline

    43 & 2022 & CLD-Net~\cite{chen2022cld} & BIBM & MiT &  Small polyp segmentation; local edge feature extraction & N/A \\ \hline

    44 & 2022 & BANet~\cite{lu2022boundary} & PRCV &  Res2Net-50 &  Attention-aware localization; residual pyramid convolution & N/A \\ \hline
    
    45 & 2023 & PolypSeg+~\cite{wu2023PolypSeg} & TCYB & ResNet50 & Adaptive scale context module; lightweight attention mechanism & \href{link}{https://github.com/szuzzb/polypsegplus} \\ \hline
    
    46 & 2023 & APCNet~\cite{yue2023attention} & TIM &  ResNet50 &  Attention-guided multi-level aggregation strategy; complementary information from different layers & N/A \\ \hline

    47 & 2023 & RA-DENet~\cite{wang2023ra} & CBM & Res2Net &  Improved reverse attention; distraction elimination & N/A \\ \hline

    48 & 2023 & EFB-Seg~\cite{su2023accurate} & Neurocom-puting  & ConvNet  & Boundary Embedding; semantic offset field learned & N/A \\ \hline
    
    49 & 2023 & PPNet~\cite{hu2023ppnet} & CBM & P2T & Channel attention; pyramid feature fusion & N/A \\ \hline

    50 & 2023 & Fu-TransHNet~\cite{wang2023cooperation} & arxiv  & HardNet68 & CNN and Transformer; multi-view learning & N/A \\ \hline

    51 & 2023 & DilatedSegNet~\newline\cite{tomar2023dilatedsegnet} & MMM  & ResNet50 & Dilated convolution pooling block; convolutional attention & \href{link}{https://github.com/suyanzhou626/FeDNet-BSPC} \\ \hline

    52 & 2023 & FeDNet~\cite{su2023fednet} & BSPC  & PVT & Decouple edge features and main body features & \href{link}{https://github.com/suyanzhou626/FeDNet-BSPC} \\ \hline

    53 & 2023 & PEFNet~\cite{nguyen2023pefnet} & MMM  & EfficientNet V2-L & Positional encoding and information fusion & \href{link}{https://github.com/huyquoctrinh/PEFNet} \\ \hline
    
    54 & 2023 & Polyp-SAM~\cite{li2023polyp} & arxiv  & ViT &  Fine-tuned SAM model for polyp segmentation~\cite{kirillov2023segment}. & \href{link}{https://github.com/ricklisz/Polyp-SAM} \\ \hline

    55 & 2023 & ESFPNet~\cite{chang2023esfpnet} & SPIE MI & MiT &  Efficient stage-wise feature pyramid decoder & \href{link}{https://github.com/dumyCq/ESFPNet} \\ \hline

    56 & 2023 & TransNetR~\cite{jha2023transnetr} & MIDL  & ResNet50 & Transformer-based residual network; multi-center out-of-distribution testing & \href{link}{https://github.com/DebeshJha/TransNetR} \\ \hline

    57 & 2023 & CASCADE~\cite{rahman2023medical} & WACV  & PVTv2/ TransUNet & Cascaded attention-based decoder; multi-stage loss optimization; feature aggregation & \href{link}{https://github.com/SLDGroup/CASCADE} \\ \hline 
    
    58 & 2023 & CFA-Net~\cite{zhou2023cross}  & PR  & Res2Net-50 & Cross-level feature aggregated; boundary aggregated & \href{link}{https://github.com/taozh2017/CFANet} \\ \hline 

    \hline
    \end{tabular}
\end{table*}

\section{Polyp Segmentation Models}
\label{sec:models}

Recent research in polyp segmentation has leveraged popular deep-learning methods to achieve remarkable results. In contrast, earlier approaches relied primarily on manually engineered features for polyp segmentation. A summary of these models can be found in Tables~\ref{tab:001} and \ref{tab:002}. To provide a comprehensive review of these polyp segmentation algorithms, we will introduce them from the following aspects.

\begin{enumerate}[1)]

\item \textbf{Traditional models}. They rely primarily on manually designed features such as color, texture, and shape information to formulate algorithms. 
\item \textbf{Deep models}. Deep learning models automatically learn deep features, which enables them to handle unstructured and complex data and provides them with more powerful expressive capabilities.
\item \textbf{Boundary-aware models}. Edge information is crucial in providing boundary cues to boost the segmentation performance, therefore we will discuss the application of edge information in some existing models. 
\item \textbf{Attention-aware models}. The analysis of the different attention strategies used in polyp segmentation research has provided insights for the potential design of attention modules in future work.
\item \textbf{Feature fusion models}. The integration and enhancement of multi-level features can significantly enhance model performance. Therefore, we investigate the effectiveness of feature fusion strategies in polyp segmentation models.

\end{enumerate}

\subsection{Traditional Models}

Early works primarily relied on manually designed features, such as color, texture, and shape, and then employed traditional machine learning techniques for heuristic modeling. For instance, Yao \etal~\cite{yao2004colonic} proposed an automatic polyp segmentation method that combines knowledge-guided intensity adjustment, fuzzy c-means clustering, and deformable models. Lu \etal~\cite{lu2008accurate} proposed a three-stage probabilistic binary classification method that integrates low-level and mid-level information to segment polyps in 3D CT colonography. Gross \etal~\cite{gross2009polyp} studied a segmentation algorithm that enhances primary edges through multiscale filtering. Ganz \etal~\cite{ganz2012automatic} utilized prior knowledge of polyp shapes using narrow band imaging (NBI) to optimize the inherent scale selection problem of gPb-OWT-UCM. This optimization aims to achieve better segmentation results by incorporating the shape information of polyps.

\subsection{Deep Models}

However, the aforementioned methods are constrained by the limited information representation capabilities of manual features. They lack generalization capabilities and are not suitable for large-scale deployment. We will review some representative deep learning-based methods in the field of polyp segmentation.

\textbf{CNN-based methods}. Thanks to the development of convolutional neural network (CNN), especially with the introduction of U-Net~\cite{ronneberger2015u}, many models inspired by this architecture have demonstrated promising results. ACSNet~\cite{zhang2020adaptive} modifies the skip connections in U-Net into local context extraction modules and adds a global information extraction module. The features are integrated and then adaptively selected based on a channel attention strategy. EU-Net~\cite{patel2021enhanced} is an enhanced U-Net framework that enhances semantic information and introduces an adaptive global context module to extract key features. MSNet~\cite{zhao2021automatic} designs a subtraction unit to generate difference features between adjacent layers and pyramidically equips it with different receptive fields to capture multi-scale information. In addition, it introduces LossNet to supervise the perceptual features at each layer. PEFNet~\cite{nguyen2023pefnet} utilizes an improved U-Net in the merging phase and embeds new location feature information. Thanks to the rich knowledge of positional information and concatenated features, this model achieves higher accuracy and universality in polyp segmentation.

\textbf{Transformer-based methods}. CNNs have limitations in capturing long-range dependencies. However, the Transformer model, which first gained popularity in natural language processing, has been introduced to computer vision. Its advantage in capturing long-range dependencies has sparked a research trend in Vision Transformers. MSRAformer~\cite{wu2022msraformer} adopts a Swin Transformer as the encoder with a pyramid structure to extract features at different stages and utilizes a multi-scale channel attention module to extract multi-scale feature information. DuAT~\cite{tang2023duat} is a dual-aggregation transformer network for polyp segmentation, which aggregates global and local spatial features and locates multi-scale objects. SSFormer~\cite{wang2022stepwise} incorporates PVTv2~\cite{wang2022pvt} and Segformer as the encoder while introducing a novel progressive local decoder. This decoder is specifically designed to complement the pyramid Transformer backbone by emphasizing local features and mitigating attention dispersion. ColonFormer~\cite{duc2022colonformer} adopts a lightweight architecture based on the transformer as the encoder and uses a hierarchical network structure for learning multi-level features in the decoder. TransNetR~\cite{jha2023transnetr} is a Transformer-based residual network, comprising a pre-trained encoder, three decoder blocks, and an upsampling layer, demonstrating excellent real-time processing speed and multi-center generalization capability. Polyp-PVT~\cite{dong2021polyp} cascades to aggregate advanced semantic and positional information, employing a similarity aggregation module to extend high-level features to the entire region.

\textbf{Hybrid methods}. Furthermore, numerous models have combined the strengths of both CNN and Transformer, capturing both local context information and long-range dependencies to significantly enhance segmentation performance. TransFuse~\cite{zhang2021transfuse} combines the Transformer and CNN branches in parallel. It employs a self-attention mechanism and a multi-modal fusion mechanism to integrate features from different branches, and uses spatial attention to enhance local details. LAPFormer~\cite{nguyen2022lapformer} employs a hierarchical transformer encoder with a CNN decoder. Additionally, it introduces a progressive feature fusion module to integrate multi-scale features and incorporates a feature refinement module and a feature selection module for feature handling. PPFormer~\cite{cai2022using} adopts a shallow CNN encoder and deep Transformer-based encoder to extract features. HSNet~\cite{zhang2022hsnet} utilizes a dual-branch structure composed of Transformer and CNN networks to capture both long-range dependencies and local appearance details. Fu-TransHNet~\cite{wang2023cooperation} designs a novel feature fusion module to fully make use of the local and global features obtained from CNN and Transformer networks.

\subsection{Boundary-aware Models}

The utilization of edge information as a guiding strategy was initially prevalent in object detection and has been increasingly applied to polyp segmentation tasks. Precise edge-aware representation aids in improving segmentation accuracy. In the following sections, we will review several models that exhibit exceptional capability in perceiving and incorporating edge information. FeDNet~\cite{su2023fednet} simultaneously optimizes the main body and edges to improve polyp segmentation performance. BSCA-Net~\cite{lin2022bsca} utilizes bit-plane slicing information to effectively extract boundary information. BoxPolyp~\cite{wei2022boxpolyp} mitigates overfitting issues by using box annotations, iteratively enhancing the segmentation model to generate fine-grained polyp regions. BDG-Net~\cite{qiu2022bdg} utilizes a boundary distribution generation module to aggregate high-level features, which are taken as supplementary spatial information and fed to the boundary distribution guided decoder for guiding polyp segmentation. ICBNet~\cite{xiao2022icbnet} enhances encoder features from preliminary segmentation and boundary prediction by incorporating context and boundary-aware information. CLD-Net~\cite{chen2022cld} employs a progressive strategy to extract edge features and addresses the issue of feature loss during downsampling by combining a local edge feature extraction module with a local-global feature fusion module. BANet~\cite{lu2022boundary} identifies the main location of polyps through an attention-aware localization module. SFA~\cite{fang2019selective} improves segmentation performance by constructing a selective feature aggregation network with regional and boundary constraints. FCBFormer~\cite{sanderson2022fcn} fully leverages the strengths of fully convolutional networks (FCNs) and transformers for polyp segmentation.

\subsection{Attention-aware Models}
Achieving superior performance in polyp segmentation requires the ability to prioritize relevant information. By incorporating an attention mechanism, this problem can be effectively addressed. This allows the model to focus on the most important features, ultimately improving segmentation performance. CASCADE~\cite{rahman2023medical} takes advantage of the multi-scale features using a hierarchical vision transformer architecture. APCNet~\cite{yue2023attention} extracts multi-level features from a pyramid structure and leverages an attention-guided multi-level aggregation strategy to enhance each layer's context features. RA-DENet~\cite{wang2023ra} enhances the representation of different regions through inverse attention, and then eliminates noise through distractor removal. TGANet~\cite{tomar2022tganet} incorporates a text attention mechanism, which utilizes features related to the size and number of polyps to adapt to varying polyp sizes and effectively handle scenarios with multiple polyps. LDNet~\cite{zhang2022lesion} extracts global context features from the input image and then updates them iteratively based on the lesion features predicted by the segmentation. SANet~\cite{wei2021shallow} eliminates the impact of colors through a color swap operation, and then filters out background noise from shallow features based on a shallow attention module.

\subsection{Feature Fusion Models}

It is crucial to incorporate multi-scale features to effectively handle variations in object size~\cite{wang2019dermoscopic}. Moreover, integrating multi-level features through feature fusion and leveraging both high-level and low-level features can significantly enhance segmentation performance~\cite{wang2019bi,wang2021knowledge}. In the context of polyp segmentation tasks, various feature fusion strategies have been utilized to achieve this goal. CFA-Net~\cite{zhou2023cross} is a novel cross-level feature aggregation network that adopts a hierarchical strategy to incorporate edge features into the dual-stream segmentation network. EFB-Seg~\cite{su2023accurate} enhances multi-level feature fusion by introducing a feature fusion module that utilizes the learned semantic offset field to align multi-level feature maps, thereby addressing the issue of feature misalignment. MSRF-Net~\cite{srivastava2021msrf} innovatively uses a dual-scale dense fusion block to exchange multi-scale features with different receptive fields. DCRNet~\cite{yin2022duplex} captures both intra-image and inter-image context relationships. Within the image, a positional attention module is presented to capture pixel-level context information. PPNet~\cite{hu2023ppnet} uses a channel attention scheme in the pyramid feature fusion module to learn global context features, thereby guiding the information transformation of the decoder branches. PraNet~\cite{fan2020pranet} aggregates high-level features using parallel partial decoders, uses a reverse attention module to mine boundary clues, and establishes the relationship between region and boundary cues. PolypSeg~\cite{zhong2020polypseg} aggregates multi-scale context information and focuses on the target area using an improved attention mechanism.

\subsection{Video Polyp Segmentation}

To facilitate the deployment of automated segmentation methods in clinical settings, a shift in focus has been observed in some studies toward video-based polyp segmentation methods. By considering temporal information, these approaches aim to overcome the limitations of single-image segmentation and enable more precise and efficient segmentation in real-time scenarios within clinical requirements. ESFPNet~\cite{chang2023esfpnet} constructs a pre-trained mixed Transformer (MiT) encoder and an efficient stage-wise feature pyramid decoder, in which the MiT uses overlapping path merging modules and self-attention prediction. PNS+~\cite{ji2022video} extracts long-term spatiotemporal representations using global encoders and local encoders and refines them gradually with normalized self-attention blocks. SSTAN~\cite{zhao2022semi} presents a semi-supervised video polyp segmentation task that only requires sparsely annotated frames for training. PNS-Net~\cite{ji2021progressively} leverages standard self-attention modules and CNN to efficiently learn representations from polyp videos in realtime without the need for post-processing. NanoNet~\cite{jha2021nanonet} has fewer parameters and can be integrated with mobile and embedded devices. It utilizes a pre-trained MobileNetV2~\cite{sandler2018mobilenetv2} as the encoder. In the architecture between the encoder and decoder, a modified residual block is incorporated to enhance the generalization capability of the decoder.

\section{Polyp Segmentation Datasets}
\label{sec:dataset}

\begin{figure}[h]
    \centering
    \includegraphics[width=1\linewidth]{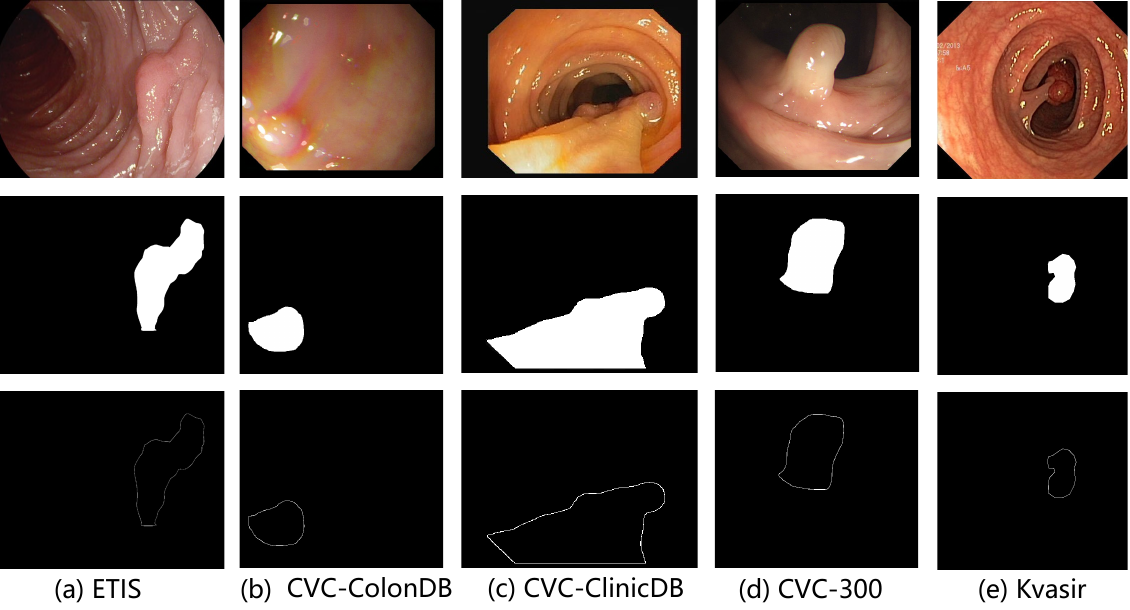} 
    \caption{Examples of images, ground truth maps, and edges in five polyp segmentation datasets, including (a) ETIS-LaribPolypDB~\cite{silva2014toward}, (b) CVC-ColonDB~\cite{tajbakhsh2015automated}, (c) CVC-ClinicDB~\cite{bernal2015wm}, (d) CVC-300~\cite{vazquez2017benchmark}, and (e) Kvasir-SEG~\cite{jha2020kvasir}. In each dataset, the image, ground truth maps, and edges are shown from top to bottom.}  \label{fig_datasetExample}
\end{figure}

The rapid progress in the field of medical image segmentation has led to the construction of various public benchmark datasets specifically designed for polyp segmentation tasks. By utilizing these benchmark datasets, researchers can compare their methods against established baselines, facilitate reproducible research, and encourage further advances in the field of polyp segmentation. Table~\ref{tab:03} summarizes eight popular image-level polyp segmentation datasets, and Fig.~\ref{fig_datasetExample} shows examples of images (including edge maps, and annotations) from these datasets. Moreover, we provide some details about each dataset below. Please note that some video-level polyp segmentation datasets are summarized in Table~\ref{tab:03}, and some works~\cite{tajbakhsh2015automated,bernal2015wm,ma2021ldpolypvideo,wang2023s,ji2022video} are based on them.

\begin{enumerate}[1)]
\item ETIS-LaribPolypDB~\cite{silva2014toward} is a collection of early colorectal polyp images consisting of $196$ polyp instances, each captured at a resolution of $966\times{1225}$ pixels. 

\item  CVC-ClinicDB~\cite{bernal2015wm} is a dataset collected from clinical cases at a hospital in Barcelona, Spain, and is developed from 23 colonoscopy video studies acquired with white light. This dataset contains 612 high-resolution color images with a resolution of $576\times{768}$, all derived from clinical colonoscopy examinations. Each image comes with a corresponding manual annotation file that clearly delineates the location of the polyp.

\item  CVC-ColonDB~\cite{tajbakhsh2015automated} is maintained by the Computer Vision Center (CVC) in Barcelona. It includes $380$ colonoscopy images with a resolution of $500\times{574}$ and segmentation masks manually annotated to precisely mark the location of polyps.

\item CVC-300~\cite{vazquez2017benchmark} includes $60$ colonoscopy images with a resolution of $500\times{574}$.

\item CVC-EndoSceneStill~\cite{vazquez2017benchmark} includes CVC-ClinicDB and CVC-300, thereby containing 912 colonoscopy examination images along with corresponding annotations.

\item Kvasir-SEG~\cite{jha2020kvasir} contains $1,000$ images of gastrointestinal polyps along with corresponding segmentation masks and bounding boxes. These were manually annotated by a doctor and validated by a gastroenterology expert.

\item PICCOLO~\cite{sanchez2020piccolo} consists of $3,433$ clinical colonoscopy images from 48 patients, including white light and narrow-band images. It also provides annotations including the number and size of polyps detected during colonoscopy. The data is divided into a training set ($2,203$), a validation set ($897$), and a test set ($333$).

\item PolypGen~\cite{ali2023multi} originates from colonoscopy detection images of over $300$ patients from six different centers. It includes both single-frame data and sequence data, containing $3,762$ annotated polyp labels. The delineation of polyp boundaries was validated by six senior experts in gastroenterology. Specifically, the dataset also contains $4,275$ negative samples.
\end{enumerate}

\renewcommand\arraystretch{1.7}
\begin{table*}[t!]
    \centering
    
    \caption{Statistics on the datasets are made according to the year (Year), the publication journal (Pub.), the dataset size (Size), the number of objects in the images (Obj.), and the resolution (Resolution). More detailed information about each dataset can be found in \secref{sec:dataset}.
    }
    \scriptsize
    \label{tab:03}
     \setlength{\tabcolsep}{3.0pt}
    \begin{tabular}{|p{0.5cm}|p{0.3cm}|p{4.3cm}|p{0.8cm}|p{1.2cm}|p{1.2cm}|p{1.5cm}|p{4.3cm}|}
        \hline
         &\# & Dataset & Year & Pub. & Size & Obj. & Resolution \\ \hline
         \multirow{8}{*}{\myrotate{Image-level}} &
         1 & ETIS-LaribPolypDB \cite{silva2014toward} &2014  & IJCARS & 196 & Multiple & $ 966\times{1225}$  \\ 
         &2 & CVC-ClinicDB \cite{bernal2015wm} &2015  & CMIG & 612 & Multiple & $576\times{768}$  \\ 
         &3 & CVC-ColonDB \cite{tajbakhsh2015automated} &2015  & TMI & 380 & One & $500\times{574}$  \\ 
         &4 & CVC-300 \cite{vazquez2017benchmark} &2017 & JHE & 60 & One & $500\times{574}$  \\ 
         &5 & CVC-EndoSceneStill \cite{vazquez2017benchmark}   &2017  & JHE & 912 & Multiple & $[500\sim 576]\times [574\sim 768]$ \\ 
         &6 & Kvasir-SEG \cite{jha2020kvasir}   &2020  & MMM & 1,000 & Multiple & $[332\sim 1920]\times[487\sim 1072]$  \\ 
         &7 & PICCOLO \cite{sanchez2020piccolo}   &2020  & AS & 3,433 & Multiple & $[480\sim 1080]\times[854\sim 1920]$\\ 
         &8 & PolypGen \cite{ali2023multi}   &2021 & SD & 8,037 & Multiple & $[288\sim 1080]\times[384\sim 1920]$\\ \hline
         \multirow{4}{*}{\myrotate{Video-level}} 
         &9 & ASU-Mayo Clinic \cite{tajbakhsh2015automated}   &2016  & TMI & 36,458 & One & $ 550 \times 688$\\ 
         &10 & CVC-ClinicVideoDB \cite{giana}   &2017  & GIANA & 11,954 & Multiple & N/A\\ 
         &11 & LDPolypVideo \cite{ma2021ldpolypvideo}   &2021 & MICCAI & 40,266 & Multiple & $480\times 560$\\ 
         &12 & SUN-SEG \cite{ji2022video}   &2022  & MIR & 158,690 & One & $[1008\sim 1080]\times[1158\sim 1240]$\\ \hline 

    \end{tabular} 
\end{table*}    

\section{Model Evaluation and Analysis}
\label{sec:evaluation}
 
\subsection{Evaluation Metrics}
\label{sec:metrics}

We briefly review several popular metrics for polyp segmentation task, \ie, intersection over union (IoU), precision-recall (PR), specificity, dice coefficient (Dice), \textit{F}-measure ($F_{\beta}$)~\cite{borji2015salient}, mean absolute error (MAE)~\cite{perazzi2012saliency}, structural measure ($S_{\alpha}$) \cite{fan2017structure}, and enhanced-alignment measure ($E_{\phi}$)~\cite{Fan2018Enhanced}.

In this survey, we focus on pixel-level performance within a single image (or video frame), with performance calculated at the frame level. TP denotes true positives, indicating the number of pixels accurately identified as foreground (or objects of interest) in the segmentation result, corresponding to actual foreground pixels in the true annotation. FP signifies false positives, representing the number of pixels incorrectly identified as foreground in the segmentation result, despite being background pixels in the true annotation. FN refers to false negatives, denoting the number of pixels inaccurately labeled as background in the segmentation result, while these pixels are foreground in the true annotation. TN stands for true negatives, signifying the number of pixels correctly recognized as background in the segmentation result, consistent with background pixels in the true annotation.

{\textbf{PR}}. Precision represents the proportion of correctly predicted positive samples by the model. Recall (also known as sensitivity) represents the proportion of true positive samples successfully detected by the model out of all actual positive samples. PR curve can be plotted by taking recall as the \textit{x}-axis and precision as the \textit{y}-axis.
\begin{equation}
    \text{Precision} = \frac{\text{TP}}{\text{TP} + \text{FP}},~\text{Recall} = \frac{\text{TP}}{\text{TP} + \text{FN}}.
\end{equation}

{\textbf{IoU}}. Intersection over union, also known as the Jaccard coefficient, is used to measure the overlap between the predicted result and the ground truth target. It is commonly used in handling imbalanced datasets.
\begin{equation}
     \text{IoU} = \frac{\text{TP}} {\text{TP} + \text{FP} + \text{FN}}.
\end{equation}

{\textbf{\textit{F}-measure}($F_{\beta}$)}. The \textit{F}-measure, also known as the \textit{F}1-score, combines precision and recall by considering their harmonic mean. It allows adjusting the weights of precision and recall to address specific problems and is effective in handling imbalanced datasets.
\begin{equation}
    F_{\beta} = (1 + \beta^2) \frac{\text{Precision} \cdot \text{Recall}}{\beta^2 \cdot \text{Precision} + \text{Recall}}.
\end{equation}

{\textbf{Dice}}. Dice coefficient is a commonly used evaluation metric for measuring the similarity between predicted segmentation results and the ground truth segmentation targets. It provides a score ranging from 0 to 1, where 1 indicates a perfect match and 0 indicates no match at all.
\begin{equation}
    \text{Dice} = \frac{2 \times \text{TP}}{2 \times \text{TP} + \text{FP} + \text{FN}}.
\end{equation}

{\textbf{\textit{S}-measure} ($S_{\alpha}$)}. Structure measure~\cite{fan2017structure} evaluates the structural similarity of segmentation results. 
\begin{equation}
    S_\alpha = \alpha \cdot S_O + (1-\alpha) \cdot S_R.
\end{equation}

In this context, $S_O$ represents the object similarity, which measures the overlap between the segmentation result and the ground truth target. $S_R$ represents the region similarity, which assesses the structural similarity between the segmentation result and the ground truth target. The parameter $\alpha$ is used to balance the weights between these two measures.

{\textbf{\textit{E}-measure} ($E_{\phi}$)}. Enhanced-measure~\cite{Fan2018Enhanced} is used to evaluate region coverage. 
\begin{equation}
    E_\phi = \frac{(1 + \phi^2) \cdot S_O \cdot S_R}{\phi^2 \cdot S_O + S_R},
\end{equation}
where the parameter $\phi$ is a balancing parameter that adjusts the weight ratio between accuracy and region coverage.

{\textbf{MAE}}. Mean absolute error (MAE) is commonly used for evaluating regression tasks.
\begin{equation}
\text{MAE} = \frac{1}{W \times H} \sum_{i=1}^{W} \sum_{j=1}^{H} \lvert S_{i,j} - G_{i,j} \rvert,
\end{equation}
where $W$ and $H$ denote the width and height of the map, $S$ represents the predicted segmentation map, and $G$ represents the ground truth segmentation map. 

{\textbf{Specificity}}. Specificity measures the model's ability to recognize negative samples.
Specifically, specificity can be calculated using Eq.~(\ref{con:inventoryflow}):
\begin{equation}
\text{Specificity} = \frac{\text{TN}}{\text{TN} + \text{FP}}. 
\label{con:inventoryflow}
\end{equation}

The value of specificity ranges between 0 and 1, with a higher value indicating a stronger ability of the model to recognize negative samples.

\subsection{Performance Comparison and Analysis}
\label{sec:comparison}
\renewcommand\arraystretch{1.3}
\setlength{\tabcolsep}{7.0pt}
\begin{table*}[t!]
    \centering
    \caption{{Benchmark results of 24 representative polyp segmentation models (18 CNN-based and 6 Transformer-based models) on five commonly used datasets in terms of SPE and SEN. The top three results are displayed in bold, italics, and underline fonts.} }
    
    \tiny
    \vspace {-0.15cm}
    \label{tab:0011}
    
    \begin{tabular}{r|p{1.6cm}<{\centering}|p{0.46cm}<{\centering}|p{0.63cm}<{\centering}|p{0.63cm}<{\centering}|p{0.63cm}<{\centering}|p{0.63cm}<{\centering}|p{0.63cm}<{\centering}|p{0.63cm}<{\centering}|p{0.63cm}<{\centering}|p{0.63cm}<{\centering}|p{0.63cm}<{\centering}}
        \hline   
    \multirow{2}{*}{\centering \scriptsize Method} & \multirow{2}{*}{\scriptsize Pub.} & \multicolumn{2}{c|}{ETIS-Larib} & \multicolumn{2}{c|}{CVC-ColonDB} & \multicolumn{2}{c|}{CVC-ClinicDB} & \multicolumn{2}{c|}{CVC-300} & \multicolumn{2}{c}{Kvasir}\\ 
    \cline{3-12}
    & & SPE & SEN & SPE & SEN & SPE & SEN & SPE & SEN & SPE & SEN  \\
    \hline     
        UNet~\cite{ronneberger2015u} & MICCAI 2015    & 0.703 & 0.484 & 0.798 & 0.525 & 0.947 & 0.835 & 0.965 & 0.768 & 0.949 & 0.857  \\ 
        UNet++~\cite{zhou2018unet++} & MICCAI 2018    & 0.727 & 0.415 & 0.828 & 0.497 & 0.927 & 0.795 & 0.957 & 0.738 & \blu{0.986} & 0.807  \\ 
        SFA~\cite{fang2019selective}  & MICCAI 2018   & 0.781 & 0.633 & 0.861 & 0.703 & 0.919 & 0.802 & 0.934 & 0.889 & 0.965 & 0.799  \\ 
        PraNet~\cite{fan2020pranet} & MICCAI 2020     & 0.805 & 0.688 & 0.874 & 0.740 & \gre{0.990} & 0.911 & 0.988 & 0.941 & 0.978 & 0.912 \\ 
        ACSNet~\cite{zhang2020adaptive} & MICCAI 2020 & 0.775 & 0.738 & 0.873 & 0.760 & 0.956 & 0.909 & 0.984 & 0.959 & 0.973 & 0.907  \\ 
        MSEG~\cite{huang2021hardnet} & arxiv 2021     & 0.844 & 0.740 & 0.912 & 0.753 & \rev{0.992} & 0.924 & 0.989 & 0.934 & \gre{0.985} & 0.900 \\ 
        EU-Net~\cite{patel2021enhanced} & CRV 2021    & 0.871 & 0.872 & 0.939 & \blu{0.851} & 0.986 & \rev{0.960} & 0.982 & \blu{0.969} & 0.974 & \rev{0.934}  \\ 
        SANet~\cite{wei2021shallow} & MICCAI 2021     & 0.943 & \blu{0.904} & 0.952 & 0.811 & 0.989 & 0.952 & 0.989 & \rev{0.971} & \blu{0.986} & 0.915 \\ 
        MSNet~\cite{zhao2021automatic} & MICCAI 2021  & 0.893 & 0.796 & 0.931 & 0.775 & 0.975 & 0.933 & 0.988 & 0.931 & 0.981 & 0.911  \\ 
        UACANet-S~\cite{kim2021uacanet} & ACM MM 2021 & 0.887 & 0.833 & 0.958 & 0.801 & \blu{0.991} & 0.942 & \blu{0.992} & 0.959 & 0.976 & 0.911  \\ 
        UACANet-L~\cite{kim2021uacanet} & ACM MM 2021 & 0.932 & 0.813 & 0.953 & 0.754 & \rev{0.992} & 0.943 & \rev{0.993} & 0.940 & 0.983 & 0.923  \\ 
        C2FNet~\cite{sun2021context} & IJCAI 2021     & 0.902 & 0.745 & 0.894 & 0.752 & 0.973 & 0.941 & 0.988 & 0.952 & 0.974 & 0.904  \\ 
        DCRNet~\cite{yin2022duplex} & ISBI 2022       & 0.756 & 0.747 & 0.884 & 0.777 & 0.959 & 0.913 & 0.972 & 0.945 & 0.973 & 0.903  \\ 
        BDG-Net~\cite{qiu2022bdg} & SPIE MI 2022      & 0.879 & 0.820 & 0.949 & 0.827 & \gre{0.990} & 0.942 & \blu{0.992} & 0.957 & 0.984 & 0.918  \\ 
        CaraNet~\cite{lou2022caranet} & SPIE MI 2022  & 0.910 & 0.812 & 0.947 & \rev{0.858} & \blu{0.991} & 0.955 & 0.976 & 0.927 & 0.982 & 0.912  \\ 
        EFA-Net~\cite{zhou2023edge} & arxiv 2023      & 0.918 & 0.866 & 0.940 & 0.820 & 0.975 & 0.934 & 0.988 & 0.950 & \rev{0.987} & 0.914  \\ 
        CFANet~\cite{zhou2023cross} & PR 2023         & 0.910 & 0.804 & 0.953 & 0.761 & \blu{0.991} & \rev{0.960} & 0.990 & 0.952 & \gre{0.985} & 0.926  \\ 
        M2SNet~\cite{zhao2023m} & arxiv 2023          & 0.893 & 0.796 & 0.931 & 0.775 & 0.975 & 0.933 & 0.988 & 0.931 & 0.981 & 0.911  \\ \hline
        HSNet~\cite{zhang2022hsnet} & CBM 2022        & \gre{0.955} & 0.868 & \blu{0.965} & 0.821 & \rev{0.992} & 0.949 & \gre{0.991} & 0.947 & \blu{0.986} & 0.913 \\ 
        DuAT~\cite{tang2023duat} & arxiv 2022         &0.941 & 0.891 & \gre{0.962} & 0.841 & \rev{0.992} & \gre{0.956} & \gre{0.991} & 0.956 & 0.984 & \blu{0.933} \\ 
        Polyp-PVT~\cite{dong2021polyp} & AIR 2023     & \rev{0.962} & \gre{0.902} & \blu{0.965} & 0.829 & \rev{0.992} & \blu{0.959} & \rev{0.993} & 0.943 & \rev{0.987} & \gre{0.928}  \\ 
        ESFPNet~\cite{chang2023esfpnet} & MI 2023     & \blu{0.961} & \rev{0.917} & 0.961 & 0.837 & \rev{0.992} & 0.940 & \gre{0.991} & \gre{0.967} & \gre{0.985} & 0.910  \\ 
        FeDNet~\cite{su2023fednet} & BSPC 2023        &0.945 & 0.893 & \rev{0.966} & \gre{0.845} & \blu{0.991} & 0.954 & \blu{0.992} & 0.950 & \rev{0.987} & 0.924  \\ 
        SAM-B~\cite{zhou2023can} & arxiv 2023         & 0.717 & 0.415 & 0.621 & 0.246 & 0.681 & 0.309 & 0.730 & 0.412 & 0.904 & 0.510  \\ 
        SAM-H~\cite{zhou2023can} & arxiv 2023         & 0.768 & 0.525 & 0.811 & 0.480 & 0.877 & 0.547 & 0.873 & 0.685 & 0.934 & 0.769  \\ 
        SAM-L~\cite{zhou2023can} & arxiv 2023         & 0.810 & 0.567 & 0.813 & 0.500 & 0.834 & 0.623 & 0.904 & 0.756 & 0.935 & 0.774  \\

    \hline

    \end{tabular}
\end{table*} 

\renewcommand\arraystretch{1.3}
\setlength{\tabcolsep}{5.0pt}
\begin{table*}[t!]
    \centering
    \caption{Benchmark results of 24 representative polyp segmentation models (18 CNN-based and 6 Transformer-based models) on five commonly used datasets in terms of Dice, IoU, and $S_{\alpha}$. }
    
    \tiny
    \vspace {-0.15cm}
    \label{tab:007}
    
    \begin{tabular}{r|p{1.6cm}<{\centering}|p{0.46cm}<{\centering}|p{0.42cm}<{\centering}|p{0.42cm}<{\centering}|p{0.42cm}<{\centering}|p{0.42cm}<{\centering}|p{0.42cm}<{\centering}|p{0.42cm}<{\centering}|p{0.42cm}<{\centering}|p{0.42cm}<{\centering}|p{0.42cm}<{\centering}|p{0.42cm}<{\centering}|p{0.42cm}<{\centering}|p{0.42cm}<{\centering}|p{0.42cm}<{\centering}|p{0.42cm}<{\centering}}
        \hline   
    \multirow{2}{*}{\centering \scriptsize Method} & \multirow{2}{*}{\scriptsize Pub.} & \multicolumn{3}{c|}{ETIS-Larib} & \multicolumn{3}{c|}{CVC-ColonDB} & \multicolumn{3}{c|}{CVC-ClinicDB} & \multicolumn{3}{c|}{CVC-300} & \multicolumn{3}{c}{Kvasir}\\ 
    \cline{3-17}
    & & Dice & IoU & $S_{\alpha}$ & Dice & IoU & $S_{\alpha}$& Dice & IoU & $S_{\alpha}$& Dice & IoU & $S_{\alpha}$& Dice & IoU & $S_{\alpha}$ \\
    \hline     
        UNet~\cite{ronneberger2015u} & MICCAI 2015 & 0.398 & 0.335 & 0.684 & 0.504 & 0.436 & 0.710 & 0.823 & 0.755 & 0.889 & 0.710 & 0.627 & 0.843 & 0.818 & 0.746 & 0.858 \\ 
        UNet++~\cite{zhou2018unet++} & MICCAI 2018 & 0.401 & 0.344 & 0.683 & 0.482 & 0.408 & 0.692 & 0.794 & 0.729 & 0.873 & 0.707 & 0.624 & 0.839 & 0.821 & 0.743 & 0.862 \\ 
        SFA~\cite{fang2019selective}  & MICCAI 2018 & 0.297 & 0.217 & 0.557 & 0.456 & 0.337 & 0.628 & 0.700 & 0.607 & 0.793 & 0.467 & 0.329 & 0.640 & 0.723 & 0.611 & 0.782 \\ 
        PraNet~\cite{fan2020pranet} & MICCAI 2020 & 0.628 & 0.567 & 0.794 & 0.712 & 0.640 & 0.820 & 0.899 & 0.849 & 0.936 & 0.871 & 0.797 & 0.925 & 0.898 & 0.840 & 0.915 \\ 
        ACSNet~\cite{zhang2020adaptive} & MICCAI 2020 & 0.578 & 0.509 & 0.754 & 0.716 & 0.649 & 0.829 & 0.882 & 0.826 & 0.927 & 0.863 & 0.787 & 0.923 & 0.898 & 0.838 & 0.920 \\ 
        MSEG~\cite{huang2021hardnet} & arxiv 2021 & 0.700 & 0.630 & 0.828 & 0.735 & 0.666 & 0.834 & 0.909 & 0.864 & 0.938 & 0.874 & 0.804 & 0.924 & 0.897 & 0.839 & 0.912 \\ 
        EU-Net~\cite{patel2021enhanced} & CRV 2021 & 0.687 & 0.609 & 0.793 & 0.756 & 0.681 & 0.831 & 0.902 & 0.846 & 0.936 & 0.837 & 0.765 & 0.904 & 0.908 & 0.854 & 0.917 \\ 
        SANet~\cite{wei2021shallow} & MICCAI 2021 & 0.750 & 0.654 & 0.849 & 0.753 & 0.670 & 0.837 & 0.916 & 0.859 & 0.939 & 0.888 & 0.815 & 0.928 & 0.904 & 0.847 & 0.915 \\ 
        MSNet~\cite{zhao2021automatic} & MICCAI 2021 & 0.723 & 0.652 & 0.845 & 0.751 & 0.671 & 0.838 & 0.918 & 0.869 & 0.946 & 0.865 & 0.799 & 0.926 & 0.905 & 0.849 & 0.923 \\ 
        UACANet-S~\cite{kim2021uacanet} & ACM MM 2021  & 0.694 & 0.615 & 0.815 & 0.783 & 0.704 & 0.847 & 0.916 & 0.870 & 0.939 & 0.902 & 0.837 & 0.934 & 0.905 & 0.852 & 0.914 \\ 
        UACANet-L~\cite{kim2021uacanet} & ACM MM 2021 & 0.766 & 0.689 & 0.859 & 0.751 & 0.678 & 0.835 & 0.926 & 0.880 & 0.942 & \blu{0.910} & \rev{0.849} & 0.938 & 0.912 & 0.859 & 0.917 \\ 
        C2FNet~\cite{sun2021context} & IJCAI 2021 & 0.699 & 0.624 & 0.827 & 0.724 & 0.650 & 0.826 & 0.919 & 0.872 & 0.941 & 0.874 & 0.801 & 0.927 & 0.886 & 0.831 & 0.905 \\ 
        DCRNet~\cite{yin2022duplex} & ISBI 2022 & 0.556 & 0.496 & 0.736 & 0.704 & 0.631 & 0.821 & 0.896 & 0.844 & 0.933 & 0.856 & 0.788 & 0.921 & 0.886 & 0.825 & 0.911 \\ 
        BDG-Net~\cite{qiu2022bdg} & SPIE MI 2022 & 0.752 & 0.681 & 0.860 & 0.797 & 0.723 & 0.870 & 0.905 & 0.857 & 0.936 & 0.902 & 0.837 & \gre{0.940} & 0.915 & 0.863 & 0.920 \\ 
        CaraNet~\cite{lou2022caranet} & SPIE MI 2022 & 0.747 & 0.672 & 0.868 & 0.773 & 0.689 & 0.853 & 0.936 & 0.887 & \blu{0.954} & 0.900 & 0.838 & \gre{0.940} & 0.916 & 0.865 & \blu{0.929} \\ 
        EFA-Net~\cite{zhou2023edge} & arxiv 2023 & 0.749 & 0.670 & 0.858 & 0.774 & 0.696 & 0.855 & 0.919 & 0.871 & 0.943 & 0.894 & 0.830 & \blu{0.941} & 0.914 & 0.861 & \blu{0.929} \\ 
        CFANet~\cite{zhou2023cross} & PR 2023 & 0.732 & 0.655 & 0.845 & 0.743 & 0.665 & 0.835 & 0.932 & 0.883 & 0.950 & 0.893 & 0.827 & 0.938 & 0.915 & 0.861 & 0.924 \\ 
        M2SNet~\cite{zhao2023m} & arxiv 2023 & 0.723 & 0.652 & 0.845 & 0.751 & 0.671 & 0.838 & 0.918 & 0.869 & 0.946 & 0.865 & 0.799 & 0.926 & 0.905 & 0.849 & 0.923 \\ \hline
        HSNet~\cite{zhang2022hsnet} & CBM 2022 & 0.808 & \gre{0.734} & 0.882 & 0.810 & \gre{0.735} & \gre{0.868} & \rev{0.948} & \blu{0.905} & \gre{0.953} & \gre{0.903} & 0.839 & 0.937 & \rev{0.926} & \rev{0.877} & \gre{0.927} \\ 
        DuAT~\cite{tang2023duat} & arxiv 2022 & \blu{0.822} & \blu{0.746} & \gre{0.889} & \blu{0.819} & \blu{0.737} & \blu{0.873} & \rev{0.948} & \rev{0.906} & \rev{0.956} & 0.901 & \gre{0.840} & \gre{0.940} & \blu{0.924} & \blu{0.876} & \blu{0.929} \\ 
        Polyp-PVT~\cite{dong2021polyp} & AIR 2023 & 0.787 & 0.706 & 0.871 & 0.808 & 0.727 & 0.865 & \blu{0.937} & \gre{0.889} & 0.949 & 0.900 & 0.833 & 0.935 & \gre{0.917} & \gre{0.864} & 0.925 \\ 
        ESFPNet~\cite{chang2023esfpnet} & MI 2023& \rev{0.823} & \rev{0.748} & \blu{0.891} & \gre{0.811} & 0.730 & 0.864 & 0.928 & 0.883 & 0.943 & 0.902 & 0.836 & 0.934 & \gre{0.917} & 0.866 & 0.923 \\ 
        FeDNet~\cite{su2023fednet} & BSPC 2023 & \gre{0.810} & 0.733 & \rev{0.892} & \rev{0.823} & \rev{0.744} & \rev{0.878} & \gre{0.930} & 0.885 & 0.949 & \rev{0.911} & \blu{0.848} & \rev{0.946} & 0.924 & \blu{0.876} & \rev{0.933} \\ 
        SAM-B~\cite{zhou2023can} & arxiv 2023 & 0.406 & 0.370 & 0.672 & 0.215 & 0.188 & 0.553 & 0.268 & 0.231 & 0.572 & 0.371 & 0.339 & 0.650 & 0.515 & 0.459 & 0.682 \\ 
        SAM-H~\cite{zhou2023can} & arxiv 2023 & 0.517 & 0.477 & 0.730 & 0.441 & 0.396 & 0.676 & 0.547 & 0.500 & 0.738 & 0.651 & 0.606 & 0.812 & 0.778 & 0.707 & 0.829 \\ 
        SAM-L~\cite{zhou2023can} & arxiv 2023 & 0.551 & 0.507 & 0.751 & 0.468 & 0.422 & 0.690 & 0.578 & 0.526 & 0.744 & 0.726 & 0.676 & 0.849 & 0.782 & 0.710 & 0.832 \\

    \hline

    \end{tabular}
\end{table*} 

\renewcommand\arraystretch{1.3}
\setlength{\tabcolsep}{5.0pt}
\begin{table*}[t!]
    \centering
    \caption{Benchmark results of 24 representative polyp segmentation models (18 CNN-based and 6 Transformer-based models) on five commonly used datasets in terms of $F_{\beta}$, $E_{\phi}$, and MAE.}
    
    \tiny
    \vspace {-0.15cm}
    \label{tab:008}
    
    \begin{tabular}{r|p{1.6cm}<{\centering}|p{0.46cm}<{\centering}|p{0.42cm}<{\centering}|p{0.42cm}<{\centering}|p{0.42cm}<{\centering}|p{0.42cm}<{\centering}|p{0.42cm}<{\centering}|p{0.42cm}<{\centering}|p{0.42cm}<{\centering}|p{0.42cm}<{\centering}|p{0.42cm}<{\centering}|p{0.42cm}<{\centering}|p{0.42cm}<{\centering}|p{0.42cm}<{\centering}|p{0.42cm}<{\centering}|p{0.42cm}<{\centering}}
        \hline   
    \multirow{2}{*}{\centering \scriptsize Method} & \multirow{2}{*}{\scriptsize Pub.} & \multicolumn{3}{c|}{ETIS-Larib} & \multicolumn{3}{c|}{CVC-ColonDB} & \multicolumn{3}{c|}{CVC-ClinicDB} & \multicolumn{3}{c|}{CVC-300} & \multicolumn{3}{c}{Kvasir}\\ 
    \cline{3-17}
    & & $F_{\beta}$ & $E_{\phi}$ & MAE & $F_{\beta}$ & $E_{\phi}$ & MAE &$F_{\beta}$ & $E_{\phi}$ & MAE &$F_{\beta}$ & $E_{\phi}$ & MAE &$F_{\beta}$ & $E_{\phi}$ & MAE \\
    \hline     
        UNet~\cite{ronneberger2015u} & MICCAI 2015& 0.366 & 0.643 & 0.036 & 0.491 & 0.692 & 0.059 & 0.811 & 0.913 & 0.019 & 0.684 & 0.848 & 0.022 & 0.794 & 0.881 & 0.055 \\ 
        UNet++~\cite{zhou2018unet++} & MICCAI 2018 & 0.390 & 0.629 & 0.035 & 0.467 & 0.680 & 0.061 & 0.785 & 0.891 & 0.022 & 0.687 & 0.834 & 0.018 & 0.808 & 0.886 & 0.048 \\ 
        SFA~\cite{fang2019selective}  & MICCAI 2018 & 0.231 & 0.531 & 0.109 & 0.366 & 0.661 & 0.094 & 0.647 & 0.840 & 0.042 & 0.341 & 0.644 & 0.065 & 0.670 & 0.834 & 0.075 \\ 
        PraNet~\cite{fan2020pranet} & MICCAI 2020 & 0.600 & 0.808 & 0.031 & 0.699 & 0.847 & 0.043 & 0.896 & 0.963 & 0.009 & 0.843 & 0.950 & 0.010 & 0.885 & 0.944 & 0.030 \\ 
        ACSNet~\cite{zhang2020adaptive} & MICCAI 2020 & 0.530 & 0.737 & 0.059 & 0.697 & 0.839 & 0.039 & 0.873 & 0.947 & 0.011 & 0.825 & 0.939 & 0.013 & 0.882 & 0.941 & 0.032 \\ 
        MSEG~\cite{huang2021hardnet} & arxiv 2021 & 0.671 & 0.855 & 0.015 & 0.724 & 0.859 & 0.038 & 0.907 & 0.961 & 0.007 & 0.852 & 0.948 & 0.009 & 0.885 & 0.942 & 0.028 \\ 
        EU-Net~\cite{patel2021enhanced} & CRV 2021 & 0.636 & 0.807 & 0.067 & 0.730 & 0.863 & 0.045 & 0.891 & 0.959 & 0.011 & 0.805 & 0.918 & 0.015 & 0.893 & 0.951 & 0.028 \\ 
        SANet~\cite{wei2021shallow} & MICCAI 2021 & 0.685 & 0.881 & 0.015 & 0.726 & 0.869 & 0.043 & 0.909 & 0.971 & 0.012 & 0.859 & 0.962 & 0.008 & 0.892 & 0.949 & 0.028 \\ 
        MSNet~\cite{zhao2021automatic} & MICCAI 2021  & 0.677 & 0.875 & 0.020 & 0.736 & 0.872 & 0.041 & 0.913 & 0.973 & \gre{0.008} & 0.848 & 0.945 & 0.010 & 0.892 & 0.947 & 0.028 \\ 
        UACANet-S~\cite{kim2021uacanet} & ACM MM 2021 & 0.650 & 0.848 & 0.023 & 0.772 & 0.894 & 0.034 & 0.917 & 0.965 & \gre{0.008} & 0.886 & \gre{0.974} & \blu{0.006} & 0.897 & 0.948 & 0.026 \\ 
        UACANet-L~\cite{kim2021uacanet} & ACM MM 2021 & 0.740 & 0.903 & \rev{0.012} & 0.746 & 0.875 & 0.039 & 0.928 & 0.974 & \rev{0.006} & \rev{0.901} & \rev{0.977} & \rev{0.005} & 0.902 & 0.955 & 0.025 \\ 
        C2FNet~\cite{sun2021context} & IJCAI 2021 & 0.668 & 0.860 & 0.022 & 0.705 & 0.854 & 0.044 & 0.906 & 0.969 & 0.009 & 0.844 & 0.949 & 0.009 & 0.869 & 0.929 & 0.036 \\ 
        DCRNet~\cite{yin2022duplex} & ISBI 2022 & 0.506 & 0.742 & 0.096 & 0.684 & 0.840 & 0.052 & 0.890 & 0.964 & 0.010 & 0.830 & 0.943 & 0.010 & 0.868 & 0.933 & 0.035 \\ 
        BDG-Net~\cite{qiu2022bdg} & SPIE MI 2022 & 0.719 & 0.901 & \gre{0.014} & 0.781 & 0.901 & 0.028 & 0.898 & 0.959 & \gre{0.008} & 0.883 & 0.969 & \rev{0.005} & 0.906 & 0.959 & 0.025 \\ 
        CaraNet~\cite{lou2022caranet} & SPIE MI 2022 & 0.709 & 0.875 & 0.017 & 0.729 & 0.880 & 0.042 & 0.931 & \blu{0.985} & \blu{0.007} & 0.887 & \rev{0.977} & \gre{0.007} & 0.909 & \blu{0.962} & \blu{0.023} \\ 
        EFA-Net~\cite{zhou2023edge} & arxiv 2023 & 0.698 & 0.872 & 0.018 & 0.753 & 0.884 & 0.036 & 0.916 & 0.972 & 0.009 & 0.878 & 0.961 & 0.009 & 0.906 & 0.955 & \gre{0.024} \\ 
        CFANet~\cite{zhou2023cross} & PR 2023 & 0.693 & 0.881 & \gre{0.014} & 0.728 & 0.869 & 0.039 & 0.924 & \gre{0.981} & 0.007 & 0.875 & 0.962 & 0.008 & 0.903 & 0.956 & \blu{0.023} \\ 
        M2SNet~\cite{zhao2023m} & arxiv 2023 & 0.677 & 0.875 & 0.020 & 0.736 & 0.872 & 0.041 & 0.913 & 0.973 & \gre{0.008} & 0.848 & 0.945 & 0.010 & 0.892 & 0.947 & 0.028 \\ \hline
        
        HSNet~\cite{zhang2022hsnet} & CBM 2022 & \gre{0.777} & 0.904 & 0.021 & 0.796 & 0.912 & 0.032 & \rev{0.951} & \rev{0.990} & \rev{0.006} & 0.887 & 0.970 & \gre{0.007} & \rev{0.918} & \gre{0.961} & \blu{0.023} \\ 
        DuAT~\cite{tang2023duat} & arxiv 2022& \rev{0.789} & \gre{0.917} & \blu{0.013} & \blu{0.805} & \rev{0.922} & \rev{0.026} & \blu{0.950} & 0.990 & \rev{0.006} & \gre{0.890} & 0.965 & \rev{0.005} & \blu{0.916} & 0.960 & \blu{0.023} \\ 
        Polyp-PVT~\cite{dong2021polyp} & AIR 2023 & 0.750 & 0.906 & \blu{0.013} & 0.795 & \gre{0.913} & 0.031 & \gre{0.936} & \blu{0.985} & \rev{0.006} & 0.884 & 0.973 & \gre{0.007} & 0.911 & 0.956 & \blu{0.023} \\ 
        ESFPNet~\cite{chang2023esfpnet} & MI 2023 & \blu{0.786} & \blu{0.930} & \rev{0.012} & \gre{0.798} & 0.908 & \gre{0.030} & 0.930 & 0.976 & \blu{0.007} & 0.882 & 0.970 & \blu{0.006} & \gre{0.913} & 0.957 & \gre{0.024} \\ 
        FeDNet~\cite{su2023fednet} & BSPC 2023 & 0.773 & \rev{0.931} & 0.016 & \rev{0.809} & \blu{0.918} & \blu{0.029} & 0.928 & 0.978 & \blu{0.007} & \blu{0.897} & \blu{0.976} & \blu{0.006} & \rev{0.918} & \rev{0.963} & \rev{0.021} \\ 
        SAM-B~\cite{zhou2023can} & arxiv 2023 & 0.404 & 0.574 & 0.035 & 0.210 & 0.412 & 0.077 & 0.259 & 0.431 & 0.092 & 0.374 & 0.563 & 0.058 & 0.509 & 0.624 & 0.104 \\ 
        SAM-H~\cite{zhou2023can} & arxiv 2023 & 0.513 & 0.658 & 0.029 & 0.434 & 0.585 & 0.056 & 0.546 & 0.676 & 0.040 & 0.653 & 0.765 & 0.020 & 0.769 & 0.828 & 0.062 \\ 
        SAM-L~\cite{zhou2023can} & arxiv 2023 & 0.544 & 0.686 & 0.030 & 0.463 & 0.607 & 0.054 & 0.563 & 0.683 & 0.057 & 0.729 & 0.824 & 0.020 & 0.773 & 0.834 & 0.061 \\

    \hline

    \end{tabular}
\end{table*}

\subsubsection{Overall Evaluation}

\begin{figure*}[t]
    \vspace {-4mm}
    \centering
    \includegraphics[width=1\linewidth]{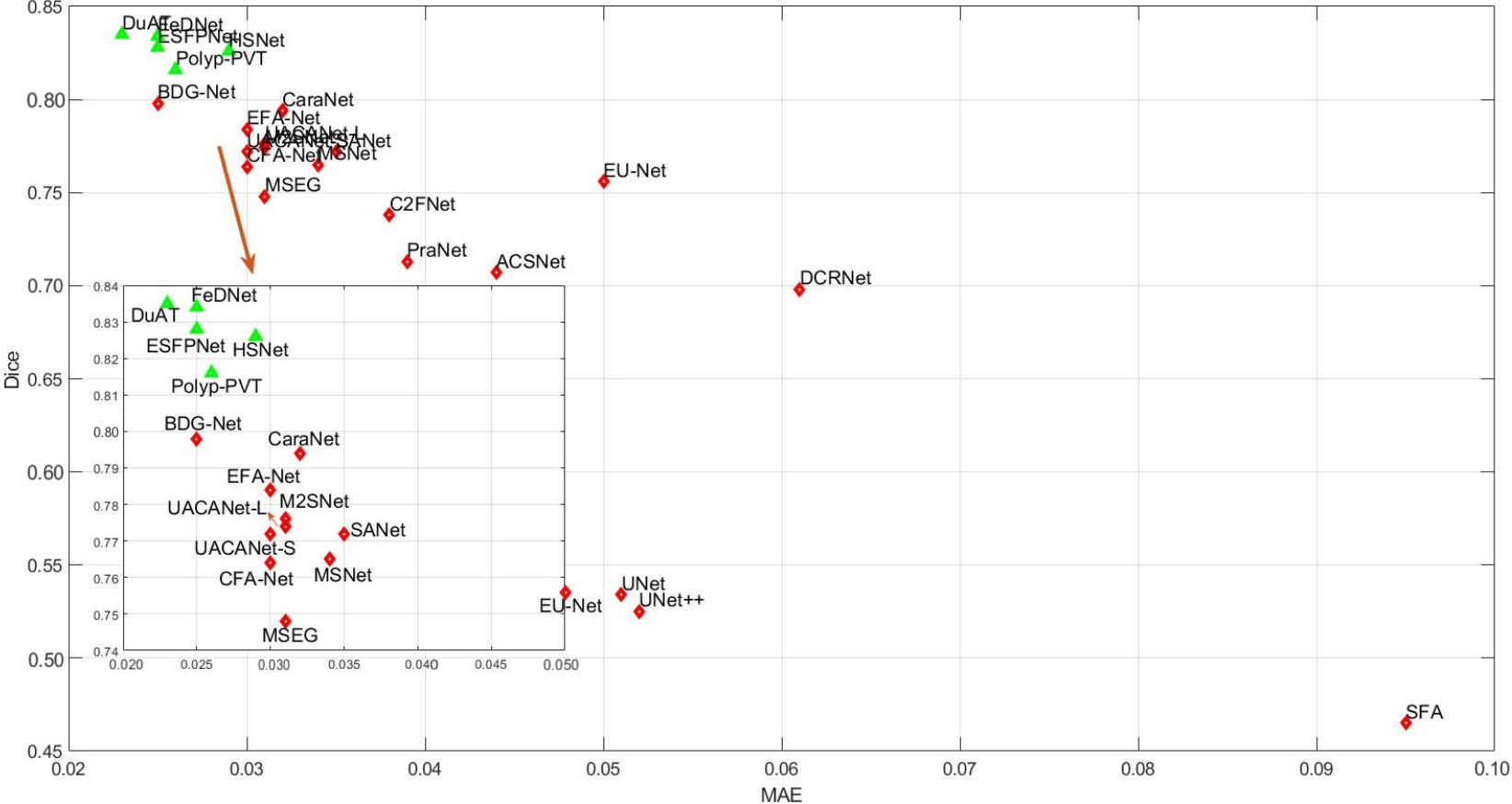} 
    \caption{A comprehensive evaluation is conducted on 23 representative deep-learning models, including UNet ~\cite{ronneberger2015u}, UNet++ ~\cite{zhou2018unet++}, SFA~\cite{fang2019selective}, PraNet~\cite{fan2020pranet}, ACSNet~\cite{zhang2020adaptive}, MSEG~\cite{huang2021hardnet}, EU-Net~\cite{patel2021enhanced}, SANet~\cite{wei2021shallow}, MSNet~\cite{zhao2021automatic}, UACANet-S~\cite{kim2021uacanet}, UACANet-L~\cite{kim2021uacanet}, C2FNet~\cite{sun2021context}, DCRNet~\cite{yin2022duplex},  BDG-Net~\cite{qiu2022bdg}, CaraNet~\cite{lou2022caranet}, EFA-Net~\cite{zhou2023edge}, CFA-Net~\cite{zhou2023cross}, M2SNet~\cite{zhao2023m}, Polyp-PVT~\cite{dong2021polyp}, HSNet~\cite{zhang2022hsnet}, DuAT~\cite{tang2023duat}, ESFPNet~\cite{chang2023esfpnet}, and FeDNet~\cite{su2023fednet}, with SAM~\cite{zhou2023can} excluded. We report the average Dice and MAE values for each model across five datasets (\ie, ETIS-LaribPolypDB~\cite{silva2014toward}, CVC-ColonDB~\cite{tajbakhsh2015automated}, CVC-ClinicDB~\cite{bernal2015wm}, CVC-300~\cite{vazquez2017benchmark}, and Kvasir-SEG~\cite{jha2020kvasir}). {Please note that the models represented in the top left corner are better,\ie, they have larger Dice scores and smaller MAE values. In this context, the green triangles represent Transformer-based models, while the red diamonds signify CNN-based models.}}\label{fig_mdice_mae}
\end{figure*}

To quantify the performance of different models, we conducted a comprehensive evaluation of 24 representative polyp segmentation models, including 1) 18 CNN-based models: UNet ~\cite{ronneberger2015u}, UNet++ ~\cite{zhou2018unet++}, SFA~\cite{fang2019selective}, PraNet~\cite{fan2020pranet}, ACSNet~\cite{zhang2020adaptive}, MSEG~\cite{huang2021hardnet}, EU-Net~\cite{patel2021enhanced}, SANet~\cite{wei2021shallow}, MSNet~\cite{zhao2021automatic}, UACANet-S~\cite{kim2021uacanet}, UACANet-L~\cite{kim2021uacanet},  C2FNet~\cite{sun2021context}, DCRNet~\cite{yin2022duplex}, BDG-Net~\cite{qiu2022bdg}, CaraNet~\cite{lou2022caranet}, EFA-Net~\cite{zhou2023edge}, CFA-Net~\cite{zhou2023cross}, and M2SNet~\cite{zhao2023m}, and 2) six Transformer-based methods: Polyp-PVT~\cite{dong2021polyp}, HSNet~\cite{zhang2022hsnet}, DuAT~\cite{tang2023duat}, ESFPNet~\cite{chang2023esfpnet}, FeDNet~\cite{su2023fednet}, and SAM-based poly segmentation model (with three different backbones, \ie, SAM-B, SAM-H, and SAM-L~\cite{zhou2023can}). First, we individually evaluated the performance of the aforementioned 24 models on SPE, SEN, Dice, IoU, $S_{\alpha}$, $F_{\beta}$, $E_{\phi}$, and MAE metrics across five public datasets (ETIS-LaribPolypDB~\cite{silva2014toward}, CVC-ColonDB~\cite{tajbakhsh2015automated}, CVC-ClinicDB~\cite{bernal2015wm}, CVC-300~\cite{vazquez2017benchmark}, and Kvasir-SEG~\cite{jha2020kvasir}). The evaluation results are presented in Table~\ref{tab:0011}, Table~\ref{tab:007} and Table~\ref{tab:008}. Note that the results of all evaluated methods have been sourced from their respective code repositories or original papers. Second, we also report the mean values of Dice and MAE across the five datasets for each model in Fig.~\ref{fig_mdice_mae}. It is worth noting that better models are shown in the upper left corner (\ie, with a larger Dice and smaller MAE). From the results shown in Fig.~\ref{fig_mdice_mae}, we have the following observations.

 \textbf{CNN vs. Transformer}. Compared with CNN-based models, Transformer-based methods achieve significantly better performance. Due to its feature extraction network structure based on the self-attention mechanism, the Transformer can effectively capture global context information. 

 \textbf{Comparison of deep models}. Among the deep learning-based models, DuAT\cite{tang2023duat}, FeDNet\cite{su2023fednet}, ESFPNet\cite{chang2023esfpnet}, Polyp-PVT\cite{dong2021polyp}, HSNet\cite{zhang2022hsnet}, and BDG-Net\cite{qiu2022bdg} achieve much better performance.

\begin{figure*}
    \centering
    \includegraphics[width=1.0\linewidth]{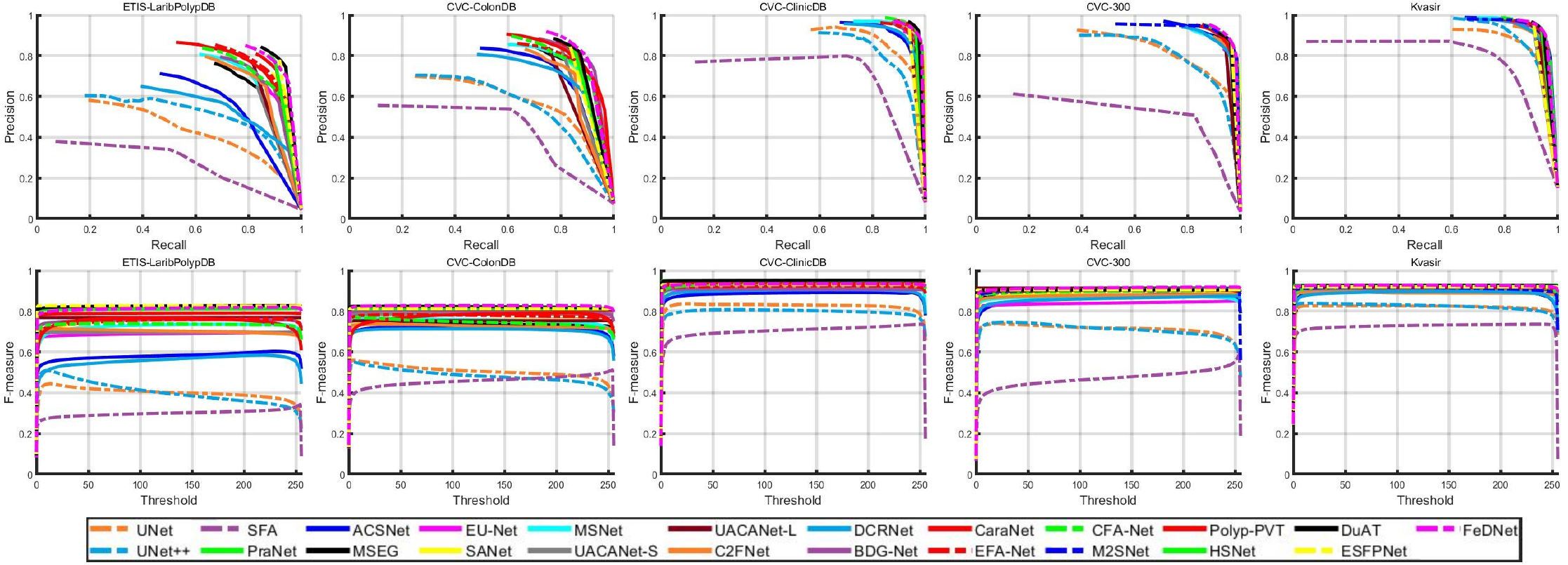} 
    \caption{The PR curves and \textit{F}-measures at different thresholds for 23 deep polyp segmentation models on five datasets from ETIS-LaribPolypDB~\cite{silva2014toward}, CVC-ColonDB~\cite{tajbakhsh2015automated}, CVC-ClinicDB~\cite{bernal2015wm}, CVC-300~\cite{vazquez2017benchmark}, and Kvasir-SEG~\cite{jha2020kvasir}.}  \label{fig_datasetresult}
\end{figure*}

Moreover, Fig.~\ref{fig_datasetresult} shows the PR and \textit{F}-measure curves for the 23 representative polyp segmentation models on five datasets (ETIS-LaribPolypDB~\cite{silva2014toward}, CVC-ColonDB~\cite{tajbakhsh2015automated}, CVC-ClinicDB~\cite{bernal2015wm}, CVC-300~\cite{vazquez2017benchmark}, and Kvasir-SEG~\cite{jha2020kvasir}).

To provide a deeper understanding of the better-performing models, we will discuss the main characteristics of the following six models in the sections below.

$\bullet$ DuAT~\cite{tang2023duat}. Dual-aggregation Transformer network uses a transformer architecture based on a pyramid structure as an encoder, and the decoder adopts a dual-stream design, building local spatial attention modules and global spatial attention modules to enhance the segmentation performance of polyps of different sizes.

$\bullet$ FeDNet~\cite{su2023fednet} decouples the input features into high-frequency edge features and low-frequency main body features through a feature decoupling operator, and then predicts by fusing the optimized features through a feature fusion operator. 

$\bullet$ ESFPNet~\cite{chang2023esfpnet} utilizes a pre-trained mixed Transformer encoder and an efficient stage-wise feature pyramid decoder structure, which fuses features from deep to shallow layers and also performs linear fusion of features from global to local, and connects them with intermediate aggregated features to obtain the final segmentation result.

$\bullet$ Polyp-PVT~\cite{dong2021polyp} introduces a pyramid vision Transformer encoder to extract multi-scale features with long-range dependencies, utilizes high-level features for side output supervision, incorporates attention mechanisms to enhance low-level features and eliminate noise, and finally fuses multi-level features.

$\bullet$ HSNet~\cite{zhang2022hsnet} is also based on a PVT encoder and suppresses noise information in low-level features by modeling the semantic spatial relationships and channel dependencies of lower-layer features. It bridges feature disparities through a semantic interaction mechanism and captures long-range dependencies and local appearance details via a dual-branch structure.

$\bullet$ BDG-Net~\cite{qiu2022bdg} aggregates high-level features to generate a boundary distribution map, which is fed into a boundary distribution-guided decoder and employs a multi-scale feature interaction strategy to enhance segmentation precision.

\subsubsection{Scale-based Evaluation}

\begin{figure*}[t]
    \vspace{-4mm}
    \centering
    \includegraphics[width=0.96\linewidth]{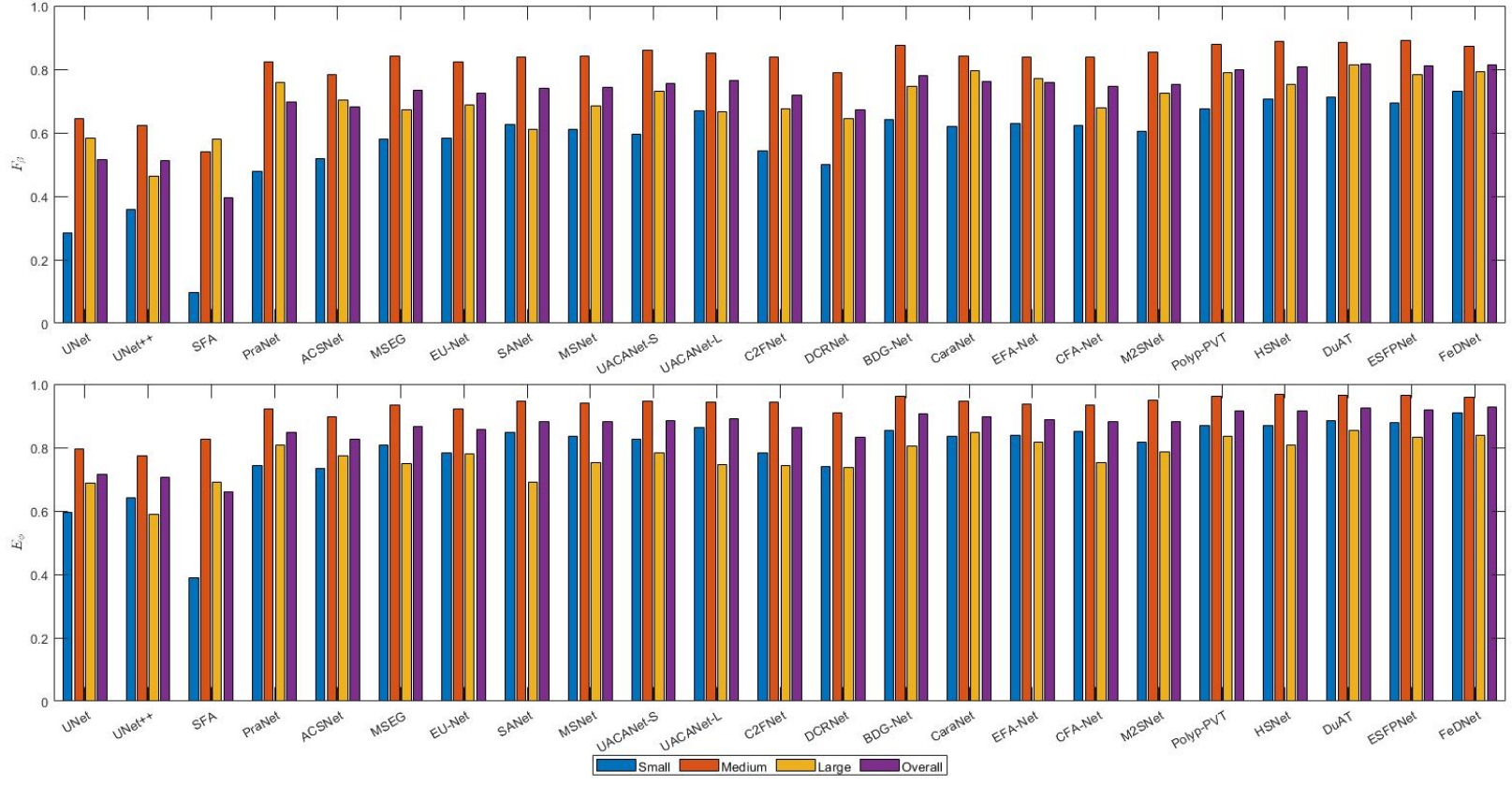} 
    \caption{{Comparison results for 26 representative polyp segmentation models are shown in terms of $F_{\beta}$ (top) and $E_{\phi}$ (bottom).}}  \label{fig_typeresult}
\end{figure*}

To investigate the influence of scale variations, we carry out evaluations on several representative polyp segmentation models. We compute the ratio ($r$) of the size of the polyp body area in a given ground truth image, which is used to characterize the size of the polyp. For this purpose, three types of polyp scales are defined: 1) when $r$ is less than $2.5\%$, the polyp is considered ``small''; 2) when $r$ is more than $20\%$, it is considered ``large''; 3) when the ratio is within the range $[2.5\%, 20\%]$, we call it ``medium''. In addition, we mix and classify the five colonoscopy datasets and derive a newly constructed dataset containing 212, 334, and 77 images of ``small'', ``medium'', and ``large'' types, respectively. Figure~\ref{fig_typeresult} presents the comparison results in relation to scale variations. Segmentation performance is represented by three metrics (\ie, IoU, $F_{\beta}$, and $E_{\phi}$).

\begin{figure}[t]
    \centering
    \includegraphics[width=1.0\linewidth]{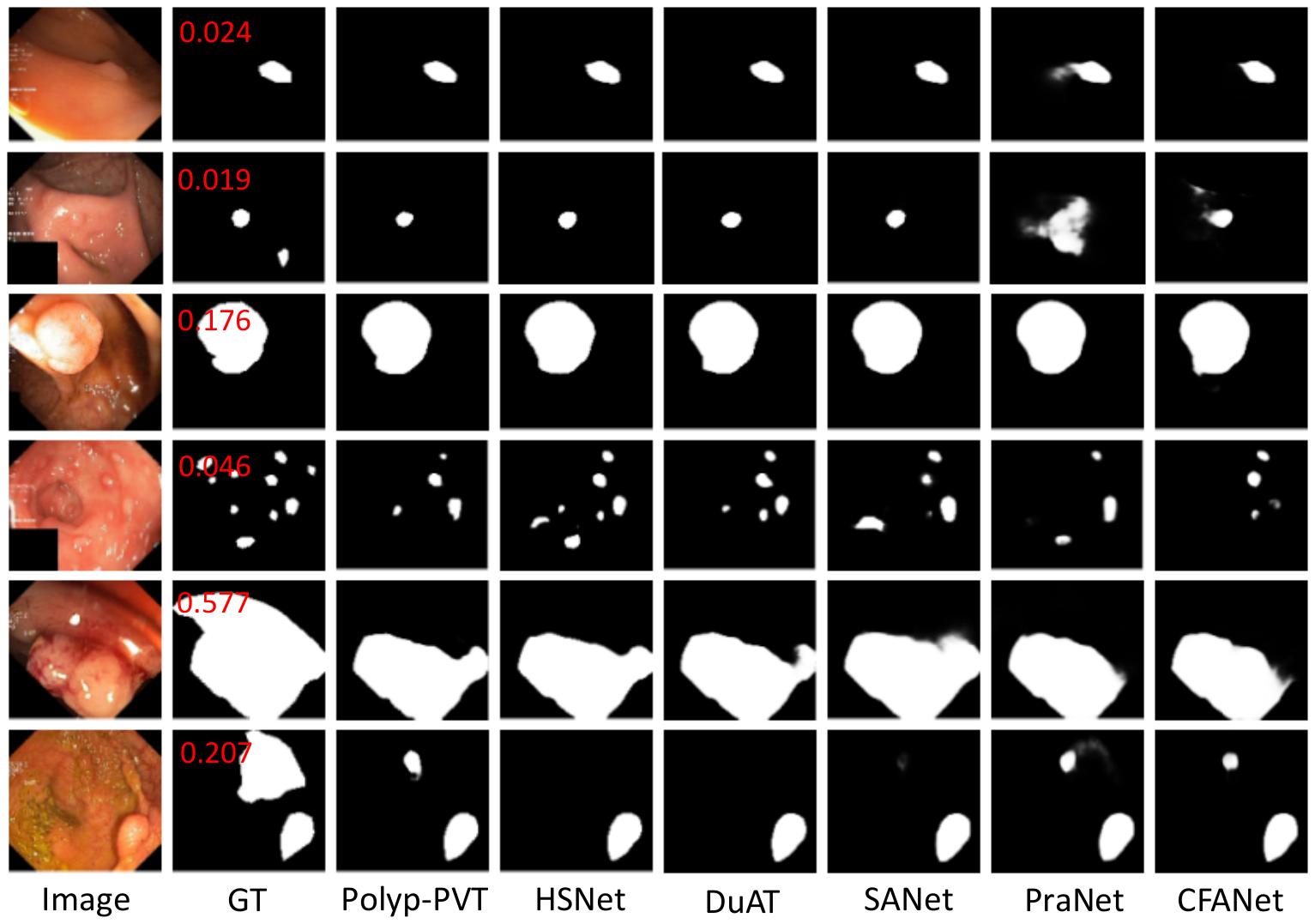} \vspace {-0.35cm}
    \caption{Some images of polyps in large, medium, and small sizes, along with the segmentation maps of six typical models, including three CNN-based models: PraNet~\cite{fan2020pranet}, SANet~\cite{wei2021shallow}, and CFA-Net~\cite{zhou2023cross}, and three transformer-based models: Polyp-PVT~\cite{dong2021polyp}, HSNet~\cite{zhang2022hsnet}, and DuAT~\cite{tang2023duat}. The numbers on the GT map represent the proportion of polyp pixels to the total number of pixels in the image. The images are sourced from the Kvasir-SEG~\cite{jha2020kvasir} dataset.}  \label{fig_mapsexample}\vspace {-0.25cm}
\end{figure}

Some sample images with different scales of polyps are shown in Fig.~\ref{fig_mapsexample}. Visual comparison results of the scale variation evaluation are shown in Table~\ref{tab:006}. According to the results, we can draw the following conclusions: (1) In terms of overall performance, DuAT~\cite{tang2023duat} and FeDNet~\cite{su2023fednet} perform better across all types, and (2) Vertical analysis reveals that most models achieve better performance in segmenting ``medium'' polyps while they exhibit relatively lower performance in segmenting other types.

\renewcommand\arraystretch{1.2}
\begin{table*}[t!]
    \centering
    
    \caption{{Performance studies based on polyp size. Comparison results for 24 representative polyp segmentation models (18 CNN-based models and 6 Transformer-based models, in this context, SAM~\cite{zhou2023can} refers to SAM-L) are provided in terms of MAE, specificity (SPE), sensitivity (SEN) and $S_{\alpha}$.}}
    
    \tiny
    \label{tab:006}
    \setlength{\tabcolsep}{4.4pt}
    \begin{tabular}{p{0.2cm}<{\centering}|p{0.65cm}<{\centering}|p{0.9cm}<{\centering}|p{0.9cm}<{\centering}|p{0.9cm}<{\centering}|p{0.9cm}<{\centering}|p{0.9cm}<{\centering}|p{0.9cm}<{\centering}|p{0.9cm}<{\centering}|p{0.9cm}<{\centering}|p{0.9cm}<{\centering}|p{0.9cm}<{\centering}|p{0.9cm}<{\centering}|p{0.9cm}<{\centering}}
        \hline
    
     \multirow{2}{*}{}    
     & & \multicolumn{12}{c|}{\textbf{ CNN-based models} }\\ \hline
     &Scale
     &UNet ~\cite{ronneberger2015u}  
     &UNet++ ~\cite{zhou2018unet++}
     &SFA ~\cite{fang2019selective}      
     &PraNet ~\cite{fan2020pranet} 
     &ACSNet ~\cite{zhang2020adaptive} 
     &MSEG ~\cite{huang2021hardnet}
     &EU-Net ~\cite{patel2021enhanced}     
     &SANet ~\cite{wei2021shallow} 
     &MSNet ~\cite{zhao2021automatic} 
     &UACANet -S~\cite{kim2021uacanet}
     &UACANet -L~\cite{kim2021uacanet}      
     &C2FNet ~\cite{sun2021context}  
    \\ \hline
   
    \multirow{4}{*}{\myrotate{MAE}}
    & Small
    &0.026 &0.015 &0.109 &0.037 &0.034 &\gre{0.014} &0.041 &\gre{0.014} &0.019 &0.020 &\blu{0.012} &0.026   \\
    & Medium 
    &0.040 &0.038 &0.062 &0.021 &0.030 &\gre{0.016} &0.035 &0.018 &0.019 &\gre{0.016} &0.017 &0.019  \\
    & Large
    &0.173 &0.215 &0.201 &0.120 &0.135 &0.138 &0.140 &0.164 &0.139 &0.120 &0.142 &0.150   \\
    & Overall
    &0.051 &0.052 &0.095 &0.039 &0.044 &0.031 &0.050 &0.035 &0.034 &\gre{0.030} &0.031 &0.038  \\ \hline 

    \multirow{4}{*}{\myrotate{SPE}}
    & Small 
    &0.618 &0.667 &0.662 &0.699 &0.746 &0.785 &0.875 &0.912 &0.847 &0.873 &0.912 &0.793  \\
    & Medium 
    & 0.888 & 0.911 & 0.940 & 0.955 & 0.915 & 0.964 & 0.954 & 0.980 & 0.975 & 0.975 & 0.974 & 0.974  \\
    & Large
    &0.856 & 0.869 & 0.966 & 0.937 & 0.942 & 0.946 & 0.919 & 0.941 & 0.918 & \blu{0.971} & 0.953 & 0.899 \\
    & Overall
    & 0.792 & 0.823 & 0.848 & 0.865 & 0.861 & 0.901 & 0.923 & 0.952 & 0.924 & 0.940 & 0.950 & 0.903 \\ \hline
    
    \multirow{4}{*}{\myrotate{SEN}}
    & Small 
    &0.377 & 0.402 & 0.575 & 0.569 & 0.680 & 0.660 & 0.806 & \blu{0.850} & 0.720 & 0.743 & 0.743 & 0.642 \\
    & Medium 
    & 0.701 & 0.649 & 0.799 & 0.857 & 0.850 & 0.865 & \rev{0.937} & \gre{0.906} & 0.880 & 0.901 & 0.856 & 0.874 \\
    & Large
    & 0.567 & 0.409 & 0.487 & 0.739 & 0.699 & 0.653 & 0.718 & 0.577 & 0.662 & 0.698 & 0.651 & 0.676 \\
    & Overall
    &0.574 & 0.535 & 0.684 & 0.744 & 0.773 & 0.769 & \blu{0.866} & 0.846 & 0.799 & 0.822 & 0.792 & 0.770 \\ \hline
    
    \multirow{4}{*}{\myrotate{$S_{\alpha}$}}
    & Small 
    &0.641 &0.679 &0.493 &0.732 &0.756 &0.786 &0.778 &0.826 &0.804 &0.780 &0.824 &0.769  \\
    & Medium 
    &0.800 &0.784 &0.736 &0.893 &0.876 &0.904 &0.885 &0.902 &0.907 &0.907 &0.897 &0.903  \\
    & Large
    &0.684 &0.590 &0.639 &0.791 &0.772 &0.743 &0.745 &0.698 &0.751 &0.768 &0.735 &0.741  \\
    & Overall
    &0.731 &0.724 &0.642 &0.826 &0.822 &0.844 &0.831 &0.851 &0.853 &0.846 &0.852 &0.837  \\ \hline     

    \end{tabular}
    
        \tiny
    \label{tab:006}
    \setlength{\tabcolsep}{4.4pt}
    \begin{tabular}{p{0.2cm}<{\centering}|p{0.65cm}<{\centering}|p{0.9cm}<{\centering}|p{0.9cm}<{\centering}|p{0.9cm}<{\centering}|p{0.9cm}<{\centering}|p{0.9cm}<{\centering}|p{0.9cm}<{\centering}|p{0.9cm}<{\centering}|p{0.9cm}<{\centering}|p{0.9cm}<{\centering}|p{0.9cm}<{\centering}|p{0.9cm}<{\centering}|p{0.9cm}<{\centering}}
        \hline
    
     \multirow{2}{*}{}    
     & & \multicolumn{6}{c|}{\textbf{ CNN-based models} } & \multicolumn{6}{c}{\textbf{  Transformer-based models}}\\ \hline
     &Scale
     &DCRNet ~\cite{yin2022duplex}
     &BDG-Net~\cite{qiu2022bdg} 
     &CaraNet ~\cite{lou2022caranet}  
     &EFA-Net~\cite{zhou2023edge}
     &CFA-Net~\cite{zhou2023cross} 
     &M2SNet ~\cite{zhao2023m}
     &Polyp-PVT~\cite{dong2021polyp}   
     &HSNet ~\cite{zhang2022hsnet} 
     &DuAT ~\cite{tang2023duat}
     &ESFPNet ~\cite{chang2023esfpnet} 
     &FeDNet ~\cite{su2023fednet}
     &SAM ~\cite{zhou2023can}
    \\ \hline
   
    \multirow{4}{*}{\myrotate{MAE}}
    & Small
    &0.068 &\rev{0.011} &0.022 &0.017 &\rev{0.011} &0.021 &0.017 &0.023 &\blu{0.012} &\gre{0.014} &\blu{0.012} &0.020 \\
    & Medium 
    &0.035 &\rev{0.013} &0.021 &0.019 &0.017 &\gre{0.016} &\blu{0.014} &\rev{0.013} &\rev{0.013} &\rev{0.012} &\gre{0.016} &0.038 \\
    & Large
    &0.152 &0.114 &0.106 &0.112 &0.138 &0.122 &\gre{0.101} &0.112 &\rev{0.094} &0.107 &\blu{0.099} &0.199  \\
    & Overall
    &0.061 &\blu{0.025} &0.032 &\gre{0.030} &\gre{0.030} &0.031 &0.026 &0.029 &\rev{0.023} &\blu{0.025} &\blu{0.025} &0.051 \\ \hline 

    \multirow{4}{*}{\myrotate{SPE}}
    & Small 
    &0.783 &0.846 &0.889 &0.875 &0.884 &0.876 &\gre{0.929} &0.915 &\blu{0.940} &0.924 &\rev{0.944} &0.732 \\
    & Medium 
    & 0.921 & \rev{0.986} & 0.973 & 0.971 & 0.979 & \gre{0.984} & 0.983 & 0.983 & \gre{0.984} & \blu{0.985} & 0.976 & 0.905 \\
    & Large
    & 0.840 & 0.925 & 0.937 & 0.955 & 0.963 & \rev{0.975} & 0.966 & 0.959 & 0.966 & \blu{0.971} & \gre{0.968} & 0.745 \\
    & Overall
    & 0.864 & 0.931 & 0.940 & 0.937 & 0.945 & 0.946 & \gre{0.963} & 0.957 & \rev{0.967} & \gre{0.963} & \blu{0.964} & 0.827 \\ \hline
    
    \multirow{4}{*}{\myrotate{SEN}}
    & Small 
    & 0.735 & 0.758 & 0.772 & 0.791 & 0.706 & 0.720 & 0.794 & 0.840 & 0.832 & \gre{0.844} & \rev{0.862} & 0.547 \\
    & Medium 
    & 0.853 & 0.899 & 0.909 & 0.894 & 0.879 & 0.889 & 0.894 & \blu{0.908} & 0.902 & 0.903 & 0.896 & 0.563 \\
    & Large
    & 0.654 & 0.735 & \blu{0.805} & 0.736 & 0.663 & 0.702 & 0.776 & 0.742 & \gre{0.786} & 0.758 & 0.776 & \rev{0.836} \\
    & Overall
   & 0.788 & 0.831 & 0.849 & 0.840 & 0.793 & 0.808 & 0.845 & 0.864 & 0.864 & \gre{0.865} & \rev{0.870} & 0.591 \\ \hline
    
    \multirow{4}{*}{\myrotate{$S_{\alpha}$}}
    & Small 
    &0.747 &0.824 &0.827 &0.825 &0.809 &0.807 &0.833 &\gre{0.849} &\blu{0.854} &0.842 &\rev{0.871} &0.737 \\
    & Medium 
    &0.878 &\rev{0.922} &0.912 &0.905 &0.902 &0.909 &0.916 &\rev{0.922} &\blu{0.921} &\rev{0.922} &\gre{0.918} &0.747 \\
    & Large
    &0.731 &0.790 &\rev{0.824} &0.801 &0.745 &0.768 &0.806 &0.782 &\blu{0.823} &0.796 &\gre{0.813} &0.696 \\
    & Overall
    &0.815 &0.873 &0.872 &0.865 &0.851 &0.857 &0.874 &\gre{0.880} &\blu{0.886} &0.879 &\rev{0.889} &0.737 \\ \hline     

    \end{tabular}
\end{table*}  

\section{Challenges and Future Trends}
\label{sec:challenge}

\subsection{Effective Network Structure} 

\textbf{Edge feature extraction}. The diverse shapes of polyps and the complex background of endoscopic images, combined with blurred edges and adherences of polyps often pose challenges for accurate segmentation. To further improve accuracy and robustness, and to leverage richer and more complex image features, some models introduce edge feature extraction modules (\cite{sanderson2022fcn,fang2019selective,lu2022boundary,xiao2022icbnet}) to utilize low-level features to assist in polyp segmentation. Most polyp segmentation models capture the edge information by utilizing low-level features extracted by backbone networks, which are then aggregated with high-level features to accomplish segmentation. However, after operations like convolution, the extracted image features, including edge and body features, become intertwined, making decoupling extremely challenging. Additionally, as the tasks of feature extraction and segmentation are not orthogonal, incorrect edge estimation can lead to error propagation. Therefore, an optional choice is to decouple the original images using traditional image processing methods. There has already been some work in the fields of object detection and other semantic segmentation domains. Shan \etal~\cite{shan2021decouple} utilized Fourier Transform to derive high and low-frequency components from images. These components are then input into two parallel branches to obtain edge and body features, which are subsequently merged to perform semantic segmentation. Cong \etal~\cite{cong2023frequency} designed a full-frequency perception module based on Octave Convolution. This module can automatically learn low and high-frequency features for coarse positioning, providing auxiliary information for segmentation. FeDNet ~\cite{su2023fednet} employs Laplacian pyramid decomposition to decouple the input features into high-frequency edge features and low-frequency body features. Following this, these two types of features undergo deep supervised optimization. The accuracy of polyp boundary annotations is contingent upon the quality of the image. In instances of substandard image quality, like blurriness or inadequate lighting, the boundary annotations may lack precision. In such scenarios, models dependent on clear boundary information could yield excessive or insufficient segmentation, resulting in reduced diagnostic accuracy. Furthermore, inaccurate boundary annotations may elevate the boundary loss function, potentially steering the model towards erroneous optimization paths, thus diminishing its generalization capacity and effectiveness.

\textbf{Dual-stream structure}. CNNs exhibit a powerful capability to extract local features, such as texture and shape. However, CNNs are relatively weaker in capturing long-distance global dependencies. In this context, the Transformer can serve as a supplement. Due to its self-attention mechanism, it possesses an excellent ability to capture global information. Therefore, introducing a dual-stream structure in existing polyp segmentation models can integrate local information captured by CNNs and global information captured by the Transformer, enhancing the performance of polyp segmentation. In addition, this structure may improve the model's capabilities to handle complex, uneven backgrounds and noise interference, thereby increasing the robustness of the model~\cite{zhang2022hsnet,zhang2021transfuse,cai2022using,nguyen2022lapformer,wang2023cooperation}. It is important to note that how to effectively merge the outputs from the two networks and balance their weights remains a problem. However, the tremendous potential and possibilities undoubtedly justify further exploration and research.

\subsection{Different Supervision Strategies}

Existing polyp segmentation models typically use fully-supervised strategies to learn features and produce segmentation results. However, annotating colonoscopy data is time-consuming and labor-intensive, especially for video data. To alleviate this problem, attention has been focused on weakly-supervised and semi-supervised learning and applying them to polyp segmentation tasks~\cite{tang2023duat, Wu_2021_ICCV}. Actually, semi-supervised or weakly supervised methods have been widely applied in the field of medical image segmentation. For example, Cho \etal\cite{cho2023anti} proposed a novel data augmentation approach for medical image segmentation that does not lose essential semantic information of the key objects. Judge \etal\cite{judge2023asymmetric} redefined the traditional per-pixel segmentation task as a contour regression problem and modeled the position uncertainty. Wei \etal\cite{wei2023consistency} introduced a semi-supervised medical image segmentation technique that first trains the segmentation model on a small number of unlabeled images, generates initial labels for them, and then introduces a consistency-based pseudo-label enhancement scheme to improve the quality of the model's predictions. Therefore, in the future, semi/weakly supervised methods can be used for image-level labeling and pseudo-annotation to improve the accuracy of polyp segmentation.

\subsection{Clinical Requirements}

\textbf{Clinical application prospects}. Advanced computer-aided diagnostic systems can greatly enhance the interpretability of medical images, assisting clinicians in making more precise diagnoses and treatment decisions. The integration of artificial intelligence (AI) into neoadjuvant chemoradiotherapy also improves the treatment outcomes and efficacy assessment in colorectal cancer. AI offers tremendous opportunities in the era of precision medicine~\cite{yin2023application}. However, the utilization of AI in healthcare is subject to stringent regulations~\cite{palaniappan2024global}, with only a few AI-based medical devices having received regulatory approval~\cite{muehlematter2021approval}. In comparison to traditional systems, the final stage of innovation in medical AI confronts numerous challenges, with deep learning models often criticized for their ``black box" issues. User skepticism also poses a barrier to the clinical implementation of AI models. Deep learning models rely heavily on ample labeled training data, highlighting the importance of engaging users consistently in algorithm enhancement and enhancing data quality as a potential new communication paradigm. 

\textbf{Dataset collection}. The shape, texture, and color of polyps can vary depending on the time and stage of the disease. Colonoscopy data from multiple centers often exhibit different morphologies. Existing polyp segmentation datasets mostly consist of images containing a single polyp, and there are not many specialized datasets with a large number of images. Only $23$ out of the 623 images in the aforementioned five datasets (ETIS-LaribPolypDB~\cite{silva2014toward}, CVC-ColonDB~\cite{tajbakhsh2015automated}, CVC-ClinicDB~\cite{bernal2015wm}, CVC-300~\cite{vazquez2017benchmark}, and Kvasir-SEG~\cite{jha2020kvasir}) contain multiple polyps. Therefore, models trained on these datasets often perform poorly on tasks involving multiple polyps. Although there are multiple public datasets available for polyp segmentation, their scale is quite limited. For instance, the largest dataset we present is PolypGen~\cite{ali2023multi}, which contains $3,762$ images. Therefore, it is necessary to develop a new large-scale polyp segmentation dataset to serve as a baseline for future research. In addition, collecting datasets for complex specialized scenarios is also a potential direction. For example, constructing multi-center datasets, multi-target datasets, and specialized datasets for small or large polyps could enhance the model's performance in different scenarios. In medical diagnostics, various imaging techniques such as computed tomography, ultrasound, and endoscopy are commonly used. However, data utilized in polyp segmentation methods often stem from a single modality~\cite{ali2024assessing}. Aggregation of multi-modal and multi-center datasets plays a critical role in overcoming the challenges associated with polyp detection and segmentation during colonoscopy examinations, making this a key area for future research.

\textbf{Cross-domain segmentation}. Deep learning-based polyp segmentation methods have achieved promising performance, but they often suffer from performance degradation when applied to unseen target domain datasets collected from different imaging devices. Therefore, it remains a challenge to apply existing polyp segmentation methods to unseen datasets. More importantly, manual annotation of new target datasets is tedious and labor-intensive, and leveraging the knowledge learned from the labeled source domains to boost the performance in the unlabeled target domain is highly demanded in clinical practice. To achieve this, Yang \etal~\cite{yang2021mutual} proposed a mutual-prototype adaptation network for cross-domain polyp segmentation, which significantly reduces the gap between the two domains to improve the segmentation performance on the target domain datasets. Thus, this direction deserves further exploration to develop more cross-domain segmentation models in this task.

\textbf{Real-time polyp segmentation}. Real-time segmentation is important for the polyp segmentation task, as it is anticipated that the segmented results can be immediately presented to the physician during the colonoscopy procedure for further decisions and treatment. However, the current deep learning-based models often require huge computation complexity, making them challenging to apply for real-time segmentation. Several real-time polyp segmentation models have been developed~\cite{feng2020ssn,jha2021nanonet, zhong2020polypseg,wichakam2018real,tomar2022transresu,wu2021precise,jha2021realtime,wu2023PolypSeg}. Indeed, the development of efficient lightweight networks for polyp segmentation without sacrificing performance is of utmost importance. However, it poses a significant challenge due to the inherent trade-off between model complexity and efficiency. Efficient lightweight networks can enable real-time segmentation, reduce computational costs, and facilitate deployment in resource-constrained clinical settings. Thus, overcoming this challenge involves finding innovative solutions that strike a balance between model complexity and performance, ultimately enhancing the practical applicability of polyp segmentation algorithms in clinical practice.

\subsection{Ethical Issues}

The specificity of medical issues often raises concerns about privacy. Moreover, there are inherent differences between data obtained from different centers. Models trained on data from a single center tend to perform worse when applied to unseen data acquired from different scanners or other centers. Thus, it becomes crucial to leverage the knowledge gained from labeled source domains to enhance the performance in unlabeled target domains. The aim is to reduce the domain shift observed in colonoscopy images obtained from multiple centers and devices. Federated learning emerges as a promising approach in this context, enabling multiple centers to collaboratively learn a shared prediction model while ensuring privacy protection. For instance, Liu \etal~\cite{liu2021feddg} presented a novel method of incidental learning in continuous frequency space, enabling diverse endpoints to utilize multi-source data distributions while addressing challenging constraints associated with data dispersion.

The widespread application of computer-assisted diagnostic systems has raised concerns regarding the potential for unintended bias in AI systems \cite{panigutti2021fairlens}. Specifically, AI-based segmentation methods heavily depend on training data, resulting in various errors within AI models \cite{thambawita2022singan}. The utilization of AI technologies in healthcare is subject to stringent regulations in the United States and the European Union \cite{muehlematter2021approval,pesapane2018artificial}, with accountability emerging as a significant issue warranting further exploration. Five primary themes influencing the credibility of medical AI include data quality, algorithmic biases, opacity, security considerations, and accountability \cite{zhang2023ethics}. Therefore, it is crucial not only to develop highly accurate and powerful segmentation models, but also to develop strategies to promote public acceptance of AI-assisted healthcare and to effectively manage and allocate responsibilities in the field of medical AI.

\section{Conclusion}
\label{sec:conclusion}

In this paper, to the best of our knowledge, we provide the first comprehensive review of the development and evaluation in the field of polyp segmentation. We first categorize models into traditional and deep, and then focus on reviewing existing deep models from various perspectives, followed by a summary of popular polyp segmentation datasets and detailed information for each dataset. Following that, we perform a comprehensive polyp size-based evaluation of 26 representative deep learning-based polyp segmentation models. To this end, we constructed a new dataset for 26 representative polyp segmentation models. Furthermore, we discuss some challenges and highlight open directions for future research. We hope this survey will spark more interest and understanding in this field.  \\ \\




\noindent \textbf{Data availability}\\
The models, benchmark datasets, and source code links we collected are all published at \href{https://github.com/taozh2017/Awesome-Polyp-Segmentation}{https://github.com/taozh2017/Awesome-Polyp-Segmentation}. To encourage future research in polyp segmentation, we will continue to collect newly published polyp segmentation models on this website.  \\ \\




\bibliographystyle{apa}
\bibliography{format_modification}

\begin{thebibliography}{}

\bibitem[\protect\astroncite{Guo et~al.}{2020}]{guo2020learn}
Guo, X., Yang, C., Liu, Y., and Yuan, Y. (2020).
\newblock Learn to threshold: Thresholdnet with confidence-guided manifold mixup for polyp segmentation.
\newblock {\em Proceedings of the IEEE Transactions on Medical Imaging}, 40(4):1134--1146.
\newblock Piscataway: IEEE.

\bibitem[\protect\astroncite{Yang et~al.}{2020}]{yang2020colon}
Yang, X., Wei, Q., Zhang, C., Zhou, K., Kong, L., and Jiang, W. (2020).
\newblock Colon polyp detection and segmentation based on improved mrcnn.
\newblock {\em Proceedings of the IEEE Transactions on Instrumentation and Measurement}, 70:1--10.
\newblock Piscataway: IEEE.

\bibitem[\protect\astroncite{Zhou et~al.}{2024}]{zhou2024uncertainty}
Zhou, T., Zhou, Y., Li, G., Chen, G., and Shen, J. (2024).
\newblock Uncertainty-aware hierarchical aggregation network for medical image segmentation.
\newblock {\em Proceedings of the IEEE Transactions on Circuits and Systems for Video Technology}, 34(8):7440--7453.
\newblock Piscataway: IEEE.

\bibitem[\protect\astroncite{Fan et~al.}{2022}]{fan2021concealed}
Fan, D., Ji, G., Cheng, M., and Shao, L. (2022).
\newblock Concealed object detection.
\newblock {\em Proceedings of the IEEE Transactions on Pattern Analysis and Machine Intelligence}, 44(10):6024--6042.
\newblock Piscataway: IEEE.

\bibitem[\protect\astroncite{Tajbakhsh et~al.}{2015}]{tajbakhsh2015automated}
Tajbakhsh, N., Gurudu, S., and Liang, J. (2015).
\newblock Automated polyp detection in colonoscopy videos using shape and context information.
\newblock {\em Proceedings of the IEEE Transactions on Medical Imaging}, 35(2):630--644.
\newblock Piscataway: IEEE.

\bibitem[\protect\astroncite{Iwahori et~al.}{2017}]{iwahori2017automatic}
Iwahori, Y., Hagi, H., Usami, H., Woodham, R., Wang, A., Bhuyan, M., et~al. (2017).
\newblock Automatic polyp detection from endoscope image using likelihood map based on edge information.
\newblock In {\em Proceedings of the International Conference on Pattern Recognition Applications and Methods}, pages 402--409. Setúbal: SciTePress.

\bibitem[\protect\astroncite{Bernal et~al.}{2015}]{bernal2015wm}
Bernal, J., S{\'a}nchez, F.~J., Fern{\'a}ndez-Esparrach, G., Gil, D., Rodr{\'\i}guez, C., and Vilari{\~n}o, F. (2015).
\newblock {WM-DOVA} maps for accurate polyp highlighting in colonoscopy: Validation vs. saliency maps from physicians.
\newblock {\em Computerized Medical Imaging and Graphics}, 43:99--111.
\newblock Elsevier.

\bibitem[\protect\astroncite{Wei et~al.}{2021}]{wei2021shallow}
Wei, J., Hu, Y., Zhang, R., Li, Z., Zhou, S.~K., and Cui, S. (2021).
\newblock Shallow attention network for polyp segmentation.
\newblock In {\em M. de Bruijne, P. C. Cattin, S. Cotin, et al. (Eds.), Proceedings of the 24th International conference on medical image computing and computer-assisted intervention}, pages 699--708. Cham:Springer.

\bibitem[\protect\astroncite{Zhao et~al.}{2021}]{zhao2021automatic}
Zhao, X., Zhang, L., and Lu, H. (2021).
\newblock Automatic polyp segmentation via multi-scale subtraction network.
\newblock In {\em L. Wang, Q. Dou, P. T. Fletcher, et al. (Eds.), Proceedings of the 25th International conference on medical image computing and computer-assisted intervention}, pages 120--130. Cham:Springer.

\bibitem[\protect\astroncite{Yang et~al.}{2021}]{yang2021mutual}
Yang, C., Guo, X., Zhu, M., Ibragimov, B., and Yuan, Y. (2021).
\newblock Mutual-prototype adaptation for cross-domain polyp segmentation.
\newblock {\em Proceedings of the IEEE Journal of Biomedical and Health Informatics}, 25(10):3886--3897.
\newblock Piscataway: IEEE.

\bibitem[\protect\astroncite{Kirillov et~al.}{2023}]{kirillov2023segment}
Kirillov, A., Mintun, E., Ravi, N., Mao, H., Rolland, C., Gustafson, L., et~al. (2023).
\newblock Segment anything.
\newblock In {\em Proceedings of the IEEE/CVF International Conference on Computer Vision}, pages 4015--4026. Piscataway: IEEE.

\bibitem[\protect\astroncite{Li et~al.}{2023}]{li2023polyp}
Li, Y., Hu, M., and Yang, X. (2023).
\newblock Polyp-{SAM}: Transfer {SAM} for polyp segmentation.
\newblock {\em arXiv preprint. arXiv:2305.00293}.

\bibitem[\protect\astroncite{Sanchez-Peralta et~al.}{2020}]{sanchez2020deep}
Sanchez-Peralta, L., Bote-Curiel, L., Picon, A., Sanchez-Margallo, F., and Pagador, J. (2020).
\newblock Deep learning to find colorectal polyps in colonoscopy: A systematic literature review.
\newblock {\em Artificial Intelligence in Medicine}, 108:101923.
\newblock Elsevier.

\bibitem[\protect\astroncite{Xiao et~al.}{2023}]{XIAO2023104791}
Xiao, H., Li, L., Liu, Q., Zhu, X., and Zhang, Q. (2023).
\newblock Transformers in medical image segmentation: A review.
\newblock {\em Biomedical Signal Processing and Control}, 84:104791.

\bibitem[\protect\astroncite{Gupta et~al.}{2023}]{gupta2023review}
Gupta, S., Sikka, G., and Malik, A. (2023).
\newblock A review on deep learning-based polyp segmentation for efficient colorectal cancer screening.
\newblock In {\em Proceedings of the 3rd International Conference on Secure Cyber Computing and Communication}, pages 501--506. Piscataway: IEEE.

\bibitem[\protect\astroncite{Qureshi et~al.}{2023}]{qureshi2023medical}
Qureshi, I., Yan, J., Abbas, Q., Shaheed, K., Riaz, A., Wahid, A., et~al. (2023).
\newblock Medical image segmentation using deep semantic-based methods: A review of techniques, applications and emerging trends.
\newblock {\em Information Fusion}, 90:316--352.
\newblock Elsevier.

\bibitem[\protect\astroncite{Chowdhary and Acharjya}{2020}]{chowdhary2020segmentation}
Chowdhary, C. and Acharjya, D. (2020).
\newblock Segmentation and feature extraction in medical imaging: a systematic review.
\newblock {\em Procedia Computer Science}, 167:26--36.
\newblock Elsevier.

\bibitem[\protect\astroncite{Thisanke et~al.}{2023}]{thisanke2023semantic}
Thisanke, H., Deshan, C., Chamith, K., Seneviratne, S., Vidanaarachchi, R., and Herath, D. (2023).
\newblock Semantic segmentation using vision transformers: A survey.
\newblock {\em Engineering Applications of Artificial Intelligence}, 126:106669.
\newblock Elsevier.

\bibitem[\protect\astroncite{Bennai et~al.}{2023}]{bennai2023multi}
Bennai, M., Guessoum, Z., Mazouzi, S., Cormier, S., and Mezghiche, M. (2023).
\newblock Multi-agent medical image segmentation: A survey.
\newblock {\em Computer Methods and Programs in Biomedicine}, 232:107444.
\newblock Elsevier.

\bibitem[\protect\astroncite{Yao et~al.}{2004}]{yao2004colonic}
Yao, J., Miller, M., Franaszek, M., and Summers, R. (2004).
\newblock Colonic polyp segmentation in ct colonography-based on fuzzy clustering and deformable models.
\newblock {\em IEEE Transactions on Medical Imaging}, 23(11):1344--1352.
\newblock Piscataway: IEEE.

\bibitem[\protect\astroncite{Lu et~al.}{2008}]{lu2008accurate}
Lu, L., Barbu, A., Wolf, M., Liang, J., Salganicoff, M., and Comaniciu, D. (2008).
\newblock Accurate polyp segmentation for {3D CT} colongraphy using multi-staged probabilistic binary learning and compositional model.
\newblock In {\em Proceedings of the IEEE Conference on Computer Vision and Pattern Recognition}, pages 1--8. Piscataway: IEEE.

\bibitem[\protect\astroncite{Gross et~al.}{2009}]{gross2009polyp}
Gross, S., Kennel, M., Stehle, T., Wulff, J., Tischendorf, J., Trautwein, C., et~al. (2009).
\newblock Polyp segmentation in {NBI} colonoscopy.
\newblock In {\em H.-P. Meinzer,T.M. Deserno,H. Handels, et al. (Eds.), Bildverarbeitung f{\"u}r die Medizin 2009: Algorithmen—Systeme—Anwendungen Proceedings des Workshops}, pages 252--256. Cham:Springer.

\bibitem[\protect\astroncite{Ganz et~al.}{2012}]{ganz2012automatic}
Ganz, M., Yang, X., and Slabaugh, G. (2012).
\newblock Automatic segmentation of polyps in colonoscopic narrow-band imaging data.
\newblock {\em Proceedings of the IEEE Transactions on Biomedical Engineering}, 59(8):2144--2151.
\newblock Piscataway: IEEE.

\bibitem[\protect\astroncite{Long et~al.}{2015}]{long2015fully}
Long, J., Shelhamer, E., and Darrell, T. (2015).
\newblock Fully convolutional networks for semantic segmentation.
\newblock In {\em Proceedings of the IEEE conference on computer vision and pattern recognition}, pages 3431--3440. Piscataway: IEEE.

\bibitem[\protect\astroncite{Ronneberger et~al.}{2015}]{ronneberger2015u}
Ronneberger, O., Fischer, P., and Brox, T. (2015).
\newblock {U-Net}: Convolutional networks for biomedical image segmentation.
\newblock In {\em D. Shen,T. Liu,T.M. Peters, et al. (Eds.),Proceedings of the 22nd International conference on medical image computing and computer-assisted intervention}, pages 234--241. Cham:Springer.

\bibitem[\protect\astroncite{V{\'a}zquez et~al.}{2017}]{vazquez2017benchmark}
V{\'a}zquez, D., Bernal, J., S{\'a}nchez, F., Fern{\'a}ndez-Esparrach, G., L{\'o}pez, A., Romero, A., et~al. (2017).
\newblock A benchmark for endoluminal scene segmentation of colonoscopy images.
\newblock {\em Journal of healthcare engineering}, 2017(1):4037190.
\newblock Wiley Online Library.

\bibitem[\protect\astroncite{Akbari et~al.}{2018}]{akbari2018polyp}
Akbari, M., Mohrekesh, M., Nasr-Esfahani, E., Soroushmehr, S., Karimi, N., Samavi, S., et~al. (2018).
\newblock Polyp segmentation in colonoscopy images using fully convolutional network.
\newblock In {\em Proceedings of the 40th Annual lnternational Conference of the IEEEEngineering in Medicine and Biology Society}, pages 69--72. Piscataway:IEEE.

\bibitem[\protect\astroncite{Zhou et~al.}{2018}]{zhou2018unet++}
Zhou, Z., Rahman~Siddiquee, M., Tajbakhsh, N., and Liang, J. (2018).
\newblock {UNet++}: A nested {UNet} architecture for medical image segmentation.
\newblock In {\em D. Stoyanov, Z. Taylor, G. Carneiro, et al. (Eds.), Proceedings of the 4th International Workshop on Deep Learning in Medical Image Analysis and Multimodal Learning for Clinical Decision Support}, pages 3--11. Cham:Springer.

\bibitem[\protect\astroncite{Fang et~al.}{2019}]{fang2019selective}
Fang, Y., Chen, C., Yuan, Y., and Tong, K. (2019).
\newblock Selective feature aggregation network with area-boundary constraints for polyp segmentation.
\newblock In {\em D. Shen, T. Liu,T.M. Peters, et al. (Eds.), Proceedings of the 22nd International conference on medical image computing and computer-assisted intervention}, pages 302--310. Cham:Springer.

\bibitem[\protect\astroncite{Fan et~al.}{2020}]{fan2020pranet}
Fan, D., Ji, G., Zhou, T., Chen, G., Fu, H., Shen, J., et~al. (2020).
\newblock {PraNet}: Parallel reverse attention network for polyp segmentation.
\newblock In {\em A.L. Martel, P.Abolmaesumi, D. Stoyanov, et al. (Eds.), Proceedings of the 23rd International conference on medical image computing and computer-assisted intervention}, pages 263--273. Cham:Springer.

\bibitem[\protect\astroncite{Dong et~al.}{2023}]{dong2021polyp}
Dong, B., Wang, W., Fan, D., Li, J., Fu, H., and Shao, L. (2023).
\newblock {Polyp-PVT}: Polyp segmentation with pyramidvision transformers.
\newblock {\em CAAI Artificial Intelligence Research}, 2:9150015.

\bibitem[\protect\astroncite{Srivastava et~al.}{2021}]{srivastava2021msrf}
Srivastava, A., Jha, D., Chanda, S., Pal, U., Johansen, H., Johansen, D., et~al. (2021).
\newblock {MSRF-Net}: a multi-scale residual fusion network for biomedical image segmentation.
\newblock {\em Proceedings of the IEEE Journal of Biomedical and Health Informatics}, 26(5):2252--2263.
\newblock Piscataway: IEEE.

\bibitem[\protect\astroncite{Tomar et~al.}{2022}]{tomar2022tganet}
Tomar, N., Jha, D., Bagci, U., and Ali, S. (2022).
\newblock {TGANet}: Text-guided attention for improved polyp segmentation.
\newblock In {\em L. Wang, Q. Dou, P. T. Fletcher, et al. (Eds.), Proceedings of the 25th International Conference on Medical Image Computing and Computer-Assisted Intervention}, pages 151--160. Cham:Springer.

\bibitem[\protect\astroncite{Wang et~al.}{2022}]{wang2022stepwise}
Wang, J., Huang, Q., Tang, F., Meng, J., Su, J., and Song, S. (2022).
\newblock Stepwise feature fusion: Local guides global.
\newblock In {\em L. Wang, Q. Dou, P. T. Fletcher, et al. (Eds.), Proceedings of the 25th International Conference on Medical Image Computing and Computer-Assisted Intervention}, pages 110--120. Cham:Springer.

\bibitem[\protect\astroncite{Rahman and Marculescu}{2023}]{rahman2023medical}
Rahman, M. and Marculescu, R. (2023).
\newblock Medical image segmentation via cascaded attention decoding.
\newblock In {\em Proceedings of the IEEE/CVF Winter Conference on Applications of Computer Vision}, pages 6222--6231. Piscataway: IEEE.

\bibitem[\protect\astroncite{Chang et~al.}{2023}]{chang2023esfpnet}
Chang, Q., Ahmad, D., Toth, J., Bascom, R., and Higgins, W. (2023).
\newblock {ESFPNet}: efficient deep learning architecture for real-time lesion segmentation in autofluorescence bronchoscopic video.
\newblock In {\em B.S.Gimi {\&A}. Krol (Eds.), Medical Imaging: Biomedical Applications in Molecular, Structural, and Functional Imaging}, pages 1--7. Bellingham: SPIE.

\bibitem[\protect\astroncite{Jha et~al.}{2019}]{jha2019resunet++}
Jha, D., Smedsrud, P., Riegler, M., Johansen, D., De~Lange, T., Halvorsen, P., et~al. (2019).
\newblock {ResUNet++}: An advanced architecture for medical image segmentation.
\newblock In {\em Proceedings of the IEEE international symposium on multimedia}, pages 225--2255. Piscataway: IEEE.

\bibitem[\protect\astroncite{Zhong et~al.}{2020}]{zhong2020polypseg}
Zhong, J., Wang, W., Wu, H., Wen, Z., and Qin, J. (2020).
\newblock Polypseg: An efficient context-aware network for polyp segmentation from colonoscopy videos.
\newblock In {\em A. L. Martel, P. Abolmaesumi, D. Stoyanov, et al. (Eds.), Proceedings of the 23rd International conference on medical image computing and computer-assisted intervention}, pages 285--294. Cham:Springer.

\bibitem[\protect\astroncite{Zhang et~al.}{2020}]{zhang2020adaptive}
Zhang, R., Li, G., Li, Z., Cui, S., Qian, D., and Yu, Y. (2020).
\newblock Adaptive context selection for polyp segmentation.
\newblock In {\em A. L. Martel, P. Abolmaesumi, D. Stoyanov, et al. (Eds.), Proceedings of the 23rd International Conference on Medical Image Computing and Computer-Assisted Intervention}, pages 253--262. Cham:Springer.

\bibitem[\protect\astroncite{Tomar et~al.}{2021}]{tomar2021ddanet}
Tomar, N., Jha, D., Ali, S., Johansen, H., Johansen, D., Riegler, M., et~al. (2021).
\newblock {DDANet}: Dual decoder attention network for automatic polyp segmentation.
\newblock In {\em A. del Bimbo, R. Cucchiara, S. Sclaroff, et al. (Eds.), Proceedings of the International Workshops and Challenges}, pages 307--314. Cham:Springer.

\bibitem[\protect\astroncite{Huang et~al.}{2021}]{huang2021hardnet}
Huang, C., Wu, H., and Lin, Y. (2021).
\newblock {HarDNet-MSEG}: A simple encoder-decoder polyp segmentation neural network that achieves over 0.9 mean dice and 86 fps.
\newblock {\em arXiv preprint. arXiv:2101.07172}.

\bibitem[\protect\astroncite{Patel et~al.}{2021}]{patel2021enhanced}
Patel, K., Bur, A., and Wang, G. (2021).
\newblock Enhanced {U-Net}: A feature enhancement network for polyp segmentation.
\newblock In {\em Proceedings of the 18th conference on robots and vision}, pages 181--188. Piscataway: IEEE.

\bibitem[\protect\astroncite{Tomar et~al.}{2022}]{tomar2022fanet}
Tomar, N., Jha, D., Riegler, M., Johansen, H., Johansen, D., Rittscher, J., et~al. (2022).
\newblock {FANet}: A feedback attention network for improved biomedical image segmentation.
\newblock {\em Proceedings of the IEEE Transactions on Neural Networks and Learning Systems}, 34(11):9375--9388.
\newblock Piscataway: IEEE.

\bibitem[\protect\astroncite{Kim et~al.}{2021}]{kim2021uacanet}
Kim, T., Lee, H., and Kim, D. (2021).
\newblock {UACANet}: Uncertainty augmented context attention for polyp segmentation.
\newblock In {\em Proceedings of the 29th ACM international conference on multimedia}, pages 2167--2175. New York: ACM.

\bibitem[\protect\astroncite{Sun et~al.}{2021}]{sun2021context}
Sun, Y., Chen, G., Zhou, T., Zhang, Y., and Liu, N. (2021).
\newblock Context-aware cross-level fusion network for camouflaged object detection.
\newblock {\em Z.-H. Zhou (Ed.), Proceedings of the 30th International Joint Conference on Artificial Intelligence}, pages 1025--1031.
\newblock Cham:Springer.

\bibitem[\protect\astroncite{Jha et~al.}{2021}]{jha2021comprehensive}
Jha, D., Smedsrud, P., Johansen, D., de~Lange, T., Johansen, H., Halvorsen, P., et~al. (2021).
\newblock A comprehensive study on colorectal polyp segmentation with {ResUNet++}, conditional random field and test-time augmentation.
\newblock {\em Proceedings of the IEEE Journal of Biomedical and Health Informatics}, 25(6):2029--2040.
\newblock Piscataway: IEEE.

\bibitem[\protect\astroncite{Zhang et~al.}{2021}]{zhang2021transfuse}
Zhang, Y., Liu, H., and Hu, Q. (2021).
\newblock Transfuse: Fusing transformers and {CNNs} for medical image segmentation.
\newblock In {\em M. de Bruijne, P. C. Cattin, S. Cotin, et al. (Eds.), Proceedings of the 24th International conference on medical image computing and computer-assisted intervention}, pages 14--24. Cham:Springer.

\bibitem[\protect\astroncite{Wu et~al.}{2021}]{wu2021multi}
Wu, L., Hu, Z., Ji, Y., Luo, P., and Zhang, S. (2021).
\newblock Multi-frame collaboration for effective endoscopic video polyp detection via spatial-temporal feature transformation.
\newblock In {\em M. de Bruijne, P. C. Cattin, S. Cotin, et al. (Eds.), Proceedings of the 24th International conference on medical image computing and computer-assisted intervention}, pages 302--312. Cham:Springer.

\bibitem[\protect\astroncite{Cheng et~al.}{2021}]{cheng2021learnable}
Cheng, M., Kong, Z., Song, G., Tian, Y., Liang, Y., and Chen, J. (2021).
\newblock Learnable oriented-derivative network for polyp segmentation.
\newblock In {\em M.de Bruijne, P.C. Cattin, S. Cotin, et al. (Eds.), Proceedings of the 24th International conference on medical image computing and computer-assisted intervention}, pages 720--730. Cham:Springer.

\bibitem[\protect\astroncite{Nguyen et~al.}{2021}]{nguyen2021ccbanet}
Nguyen, T., Nguyen, T., Diep, G., Tran-Dinh, A., Nguyen, T., and Tran, M. (2021).
\newblock {CCBANet}: cascading context and balancing attention for polyp segmentation.
\newblock In {\em M. de Bruijne, P. C. Cattin, S. Cotin, et al. (Eds.), Proceedings of the 24th International conference on medical image computing and computer-assisted intervention}, pages 633--643. Cham:Springer.

\bibitem[\protect\astroncite{Shen et~al.}{2021}]{shen2021hrenet}
Shen, Y., Jia, X., and Meng, M. (2021).
\newblock Hrenet: A hard region enhancement network for polyp segmentation.
\newblock In {\em M. de Bruijne, P. C. Cattin, S. Cotin, et al. (Eds.), Proceedings of the 24th International conference on medical image computing and computer-assisted intervention}, pages 559--568. Cham:Springer.

\bibitem[\protect\astroncite{Srivastava et~al.}{2022}]{srivastava2022gmsrf}
Srivastava, A., Chanda, S., Jha, D., Pal, U., and Ali, S. (2022).
\newblock {GMSRF-Net}: An improved generalizability with global multi-scale residual fusion network for polyp segmentation.
\newblock In {\em Proceedings of the 26th International Conference on Pattern Recognition}, pages 4321--4327. Piscataway: IEEE.

\bibitem[\protect\astroncite{Lou et~al.}{2022}]{lou2022caranet}
Lou, A., Guan, S., Ko, H., and Loew, M. (2022).
\newblock {CaraNet}: Context axial reverse attention network for segmentation of small medical objects.
\newblock In {\em Proceedings of the Medical Imaging 2022: Image Processing}, pages 81--92. Bellingham: SPIE.

\bibitem[\protect\astroncite{Wu et~al.}{2022}]{wu2022msraformer}
Wu, C., Long, C., Li, S., Yang, J., Jiang, F., and Zhou, R. (2022).
\newblock {MSRAformer}: Multiscale spatial reverse attention network for polyp segmentation.
\newblock {\em Computers in Biology and Medicine}, 151:106274.
\newblock Cham:Springer.

\bibitem[\protect\astroncite{Zhang et~al.}{2022}]{zhang2022hsnet}
Zhang, W., Fu, C., Zheng, Y., Zhang, F., Zhao, Y., and Sham, C. (2022).
\newblock {HSNet}: A hybrid semantic network for polyp segmentation.
\newblock {\em Computers in Biology and Medicine}, 150:106173.

\bibitem[\protect\astroncite{Patel et~al.}{2022}]{patel2022fuzzynet}
Patel, K., Li, F., and Wang, G. (2022).
\newblock Fuzzynet: A fuzzy attention module for polyp segmentation.
\newblock In {\em Proceedings of the 36th international conference on neural information processing systems workshops}, pages 1--11. Red Hook: Curran Associates.

\bibitem[\protect\astroncite{Zhang et~al.}{2022}]{zhang2022lesion}
Zhang, R., Lai, P., Wan, X., Fan, D., Gao, F., Wu, X., et~al. (2022).
\newblock Lesion-aware dynamic kernel for polyp segmentation.
\newblock In {\em L. Wang, Q. Dou, P. T. Fletcher, et al. (Eds.), Proceedings of the 25th International conference on medical image computing and computer-assisted intervention}, pages 99--109. Cham:Springer.

\bibitem[\protect\astroncite{Liao et~al.}{2022}]{liao2022hardnet}
Liao, T., Yang, C., Lo, Y., Lai, K., Shen, P., and Lin, Y. (2022).
\newblock {HarDNet-DFUS}: An enhanced harmonically-connected network for diabetic foot ulcer image segmentation and colonoscopy polyp segmentation.
\newblock {\em arXiv preprint. arXiv:2209.07313}.

\bibitem[\protect\astroncite{Qiu et~al.}{2022}]{qiu2022bdg}
Qiu, Z., Wang, Z., Zhang, M., Xu, Z., Fan, J., and Xu, L. (2022).
\newblock {BDG-Net}: boundary distribution guided network for accurate polyp segmentation.
\newblock In {\em O. Colliot, I. Isgum, B. A. Landman, et al. (Eds.), Medical Imaging: Image Processing}, pages 792--799. Bellingham: SPIE.

\bibitem[\protect\astroncite{Duc et~al.}{2022}]{duc2022colonformer}
Duc, N., Oanh, N., Thuy, N., Triet, T., and Dinh, V. (2022).
\newblock {ColonFormer}: An efficient transformer based method for colon polyp segmentation.
\newblock {\em IEEE Access}, 10:80575--80586.
\newblock Piscataway: IEEE.

\bibitem[\protect\astroncite{Sanderson and Matuszewski}{2022}]{sanderson2022fcn}
Sanderson, E. and Matuszewski, B. (2022).
\newblock {FCN-transformer} feature fusion for polyp segmentation.
\newblock In {\em Proceedings of the 26th Annual conference on medical image understanding and analysis}, pages 892--907. Cham:Springer.

\bibitem[\protect\astroncite{Yin et~al.}{2022}]{yin2022duplex}
Yin, Z., Liang, K., Ma, Z., and Guo, J. (2022).
\newblock Duplex contextual relation network for polyp segmentation.
\newblock In {\em Proceedings of the IEEE 19th international symposium on biomedical imaging}, pages 1--5. Piscataway: IEEE.

\bibitem[\protect\astroncite{Tang et~al.}{2023}]{tang2023duat}
Tang, F., Xu, Z., Huang, Q., Wang, J., Hou, X., Su, J., et~al. (2023).
\newblock {DuAT}: Dual-aggregation transformer network for medical image segmentation.
\newblock In {\em Q. Liu, H. Wang, Z. Ma, et al. (Eds.), Proceedings of the 6th Chinese Conference on Pattern Recognition and Computer Vision}, pages 343--356. Cham:Springer.

\bibitem[\protect\astroncite{Nguyen et~al.}{2022}]{nguyen2022lapformer}
Nguyen, M., Bui, T., Van~Nguyen, Q., Nguyen, T., and Van~Pham, T. (2022).
\newblock {LAPFormer}: A light and accurate polyp segmentation transformer.
\newblock {\em arXiv preprint. arXiv:2210.04393}.

\bibitem[\protect\astroncite{Cai et~al.}{2022}]{cai2022using}
Cai, L., Wu, M., Chen, L., Bai, W., Yang, M., Lyu, S., et~al. (2022).
\newblock Using guided self-attention with local information for polyp segmentation.
\newblock In {\em L. Wang, Q. Dou, P.T. Fletcher, et al. (Eds.), Proceedings of the 25th International conference on medical image computing and computer-assisted intervention}, pages 629--638. Cham:Springer.

\bibitem[\protect\astroncite{Lin et~al.}{2022}]{lin2022bsca}
Lin, Y., Wu, J., Xiao, G., Guo, J., Chen, G., and Ma, J. (2022).
\newblock {BSCA-Net}: Bit slicing context attention network for polyp segmentation.
\newblock {\em Pattern Recognition}, 132:108917.
\newblock Elsevier.

\bibitem[\protect\astroncite{Wei et~al.}{2022}]{wei2022boxpolyp}
Wei, J., Hu, Y., Li, G., Cui, S., Kevin~Zhou, S., and Li, Z. (2022).
\newblock {BoxPolyp}: Boost generalized polyp segmentation using extra coarse bounding box annotations.
\newblock In {\em L. Wang, Q. Dou, P. T. Fletcher, et al. (Eds.), Proceedings of the 25th International conference on medical image computing and computer-assisted intervention}, pages 67--77. Cham:Springer.

\bibitem[\protect\astroncite{Xiao et~al.}{2022}]{xiao2022icbnet}
Xiao, Y., Chen, Z., Wan, L., Yu, L., and Zhu, L. (2022).
\newblock {ICBNet}: Iterative context-boundary feedback network for polyp segmentation.
\newblock In {\em Proceedings of the IEEE International Conference on Bioinformatics and Biomedicine}, pages 1297--1304.

\bibitem[\protect\astroncite{Chen et~al.}{2022}]{chen2022cld}
Chen, R., Wang, X., Jin, B., Tu, J., Zhu, F., and Li, Y. (2022).
\newblock {CLD-Net}: Complement local detail for medical small-object segmentation.
\newblock In {\em Proceedings of the IEEE International Conference on Bioinformatics and Biomedicine}, pages 942--947. Piscataway: IEEE.

\bibitem[\protect\astroncite{Lu et~al.}{2022}]{lu2022boundary}
Lu, L., Zhou, X., Chen, S., Chen, Z., Yu, J., Tang, H., et~al. (2022).
\newblock Boundary-aware polyp segmentation network.
\newblock In {\em S. Yu, Z. Zhang, P. C. Yuen, et al. (Eds.), Proceedings of the 5th Chinese Conference on Pattern Recognition and Computer Vision}, pages 66--77. Cham:Springer.

\bibitem[\protect\astroncite{Wu et~al.}{2023}]{wu2023PolypSeg}
Wu, H., Zhao, Z., Zhong, J., Wang, W., Wen, Z., and Qin, J. (2023).
\newblock {PolypSeg+}: A lightweight context-aware network for real-time polyp segmentation.
\newblock {\em IEEE Transactions on Cybernetics}, 53(4):2610--2621.
\newblock Piscataway: IEEE.

\bibitem[\protect\astroncite{Yue et~al.}{2023}]{yue2023attention}
Yue, G., Li, S., Cong, R., Zhou, T., Lei, B., and Wang, T. (2023).
\newblock Attention-guided pyramid context network for polyp segmentation in colonoscopy images.
\newblock {\em Proceedings of the IEEE Transactions on Instrumentation and Measurement}, 72:1--13.
\newblock Piscataway: IEEE.

\bibitem[\protect\astroncite{Wang et~al.}{2023}]{wang2023ra}
Wang, K., Liu, L., Fu, X., Liu, L., and Peng, W. (2023).
\newblock {RA-DENet}: Reverse attention and distractions elimination network for polyp segmentation.
\newblock {\em Computers in Biology and Medicine}, 155:106704.
\newblock Elsevier.

\bibitem[\protect\astroncite{Su et~al.}{2023}]{su2023accurate}
Su, Y., Cheng, J., Zhong, C., Jiang, C., Ye, J., and He, J. (2023).
\newblock Accurate polyp segmentation through enhancing feature fusion and boosting boundary performance.
\newblock {\em Neurocomputing}, 545:126233.
\newblock Elsevier.

\bibitem[\protect\astroncite{Hu et~al.}{2023}]{hu2023ppnet}
Hu, K., Chen, W., Sun, Y., Hu, X., Zhou, Q., and Zheng, Z. (2023).
\newblock {PPNet}: Pyramid pooling based network for polyp segmentation.
\newblock {\em Computers in Biology and Medicine}, 160:107028.
\newblock Elsevier.

\bibitem[\protect\astroncite{Wang et~al.}{2023}]{wang2023cooperation}
Wang, Y., Deng, Z., Lou, Q., Hu, S., Choi, K., and Wang, S. (2023).
\newblock Cooperation learning enhanced colonic polyp segmentation based on transformer-cnn fusion.
\newblock {\em arXiv preprint. arXiv:2301.06892}.

\bibitem[\protect\astroncite{Tomar et~al.}{2023}]{tomar2023dilatedsegnet}
Tomar, N., Jha, D., and Bagci, U. (2023).
\newblock {DilatedSegNet}: A deep dilated segmentation network for polyp segmentation.
\newblock In {\em D.-T. Dang-Nguyen, C. Gurrin, M. A. Larson, et al. (Eds.), Proceedings of the 29th International conference on multimedia modeling}, pages 334--344. Cham:Springer.

\bibitem[\protect\astroncite{Su et~al.}{2023}]{su2023fednet}
Su, Y., Cheng, J., Zhong, C., Zhang, Y., Ye, J., He, J., et~al. (2023).
\newblock {FeDNet}: Feature decoupled network for polyp segmentation from endoscopy images.
\newblock {\em Biomedical Signal Processing and Control}, 83:104699.
\newblock Elsevier.

\bibitem[\protect\astroncite{Nguyen-Mau et~al.}{2023}]{nguyen2023pefnet}
Nguyen-Mau, T., Trinh, Q., Bui, N., Thi, P., Nguyen, M., Cao, X., et~al. (2023).
\newblock {PEFNet}: Positional embedding feature for polyp segmentation.
\newblock In {\em D.-T. Dang-Nguyen, C. Gurrin, M. A. Larson, et al. (Eds.), Proceedings of the 29th International Conference on Multimedia Modeling}, pages 240--251. Cham:Springer.

\bibitem[\protect\astroncite{Jha et~al.}{2024}]{jha2023transnetr}
Jha, D., Tomar, N., Sharma, V., and Bagci, U. (2024).
\newblock {TransNetR}: Transformer-based residual network for polyp segmentation with multi-center out-of-distribution testing.
\newblock {\em I. Oguz, J. H. Noble, X. Li, et al. (Eds.), Medical Imaging with Deep Learning Medical Imaging with Deep Learning}, pages 1372--1384.
\newblock Retrieved December 1, 2024, from \url{https://proceedings.mlr.press/v227/jha24a.html}.

\bibitem[\protect\astroncite{Zhou et~al.}{2023}]{zhou2023cross}
Zhou, T., Zhou, Y., He, K., Gong, C., Yang, J., Fu, H., et~al. (2023).
\newblock Cross-level feature aggregation network for polyp segmentation.
\newblock {\em Pattern Recognition}, 140:109555.
\newblock Elsevier.

\bibitem[\protect\astroncite{Wang et~al.}{2022}]{wang2022pvt}
Wang, W., Xie, E., Li, X., Fan, D., Song, K., Liang, D., et~al. (2022).
\newblock {PVT} v2: Improved baselines with pyramid vision transformer.
\newblock {\em Computational Visual Media}, 8(3):415--424.
\newblock Cham:Springer.

\bibitem[\protect\astroncite{Wang et~al.}{2019}]{wang2019dermoscopic}
Wang, X., Ding, H., and Jiang, X. (2019).
\newblock Dermoscopic image segmentation through the enhanced high-level parsing and class weighted loss.
\newblock In {\em Proceedings of the IEEE International Conference on Image Processing}, pages 245--249. Piscataway: IEEE.

\bibitem[\protect\astroncite{Wang et~al.}{2019}]{wang2019bi}
Wang, X., Jiang, X., Ding, H., and Liu, J. (2019).
\newblock Bi-directional dermoscopic feature learning and multi-scale consistent decision fusion for skin lesion segmentation.
\newblock {\em Proceedings of the IEEE Transactions on Image Processing}, 29:3039--3051.
\newblock Piscataway: IEEE.

\bibitem[\protect\astroncite{Wang et~al.}{2021}]{wang2021knowledge}
Wang, X., Jiang, X., Ding, H., Zhao, Y., and Liu, J. (2021).
\newblock Knowledge-aware deep framework for collaborative skin lesion segmentation and melanoma recognition.
\newblock {\em Pattern Recognition}, 120:108075.
\newblock Elsevier.

\bibitem[\protect\astroncite{Ji et~al.}{2022}]{ji2022video}
Ji, G., Xiao, G., Chou, Y., Fan, D., Zhao, K., Chen, G., et~al. (2022).
\newblock Video polyp segmentation: A deep learning perspective.
\newblock {\em Machine Intelligence Research}, 19(6):531--549.
\newblock Cham:Springer.

\bibitem[\protect\astroncite{Zhao et~al.}{2022}]{zhao2022semi}
Zhao, X., Wu, Z., Tan, S., Fan, D., Li, Z., Wan, X., et~al. (2022).
\newblock Semi-supervised spatial temporal attention network for video polyp segmentation.
\newblock In {\em L. Wang, Q. Dou, P. T. Fletcher, et al. (Eds.), Proceedings of the 25th International conference on medical image computing and computer-assisted intervention}, pages 456--466. Cham:Springer.

\bibitem[\protect\astroncite{Ji et~al.}{2021}]{ji2021progressively}
Ji, G., Chou, Y., Fan, D., Chen, G., Fu, H., Jha, D., et~al. (2021).
\newblock Progressively normalized self-attention network for video polyp segmentation.
\newblock In {\em M. de Bruijne, P. C. Cattin, S. Cotin, et al. (Eds.), Proceedings of the 24th International conference on medical image computing and computer-assisted intervention}, pages 142--152. Cham:Springer.

\bibitem[\protect\astroncite{Jha et~al.}{2021}]{jha2021nanonet}
Jha, D., Tomar, N., Ali, S., Riegler, M., Johansen, H., Johansen, D., et~al. (2021).
\newblock Nanonet: Real-time polyp segmentation in video capsule endoscopy and colonoscopy.
\newblock In {\em Proceedings of the IEEE 34th International Symposium on Computer-Based Medical Systems}, pages 37--43. Piscataway: IEEE.

\bibitem[\protect\astroncite{Sandler et~al.}{2018}]{sandler2018mobilenetv2}
Sandler, M., Howard, A., Zhu, M., Zhmoginov, A., and Chen, L. (2018).
\newblock Mobilenetv2: Inverted residuals and linear bottlenecks.
\newblock In {\em Proceedings of the IEEE conference on computer vision and pattern recognition}, pages 4510--4520. Piscataway: IEEE.

\bibitem[\protect\astroncite{Ma et~al.}{2021}]{ma2021ldpolypvideo}
Ma, Y., Chen, X., Cheng, K., Li, Y., and Sun, B. (2021).
\newblock {LDPolypVideo} benchmark: a large-scale colonoscopy video dataset of diverse polyps.
\newblock In {\em M. de Bruijne, P. C. Cattin, S. Cotin, et al. (Eds.), Proceedings of the 24th International conference on medical image computing and computer-assisted intervention}, pages 387--396. Cham:Springer.

\bibitem[\protect\astroncite{Wang et~al.}{2023}]{wang2023s}
Wang, A., Xu, M., Zhang, Y., Islam, M., and Ren, H. (2023).
\newblock {S$^{2}$ME}: Spatial-spectral mutual teaching and ensemble learning for scribble-supervised polyp segmentation.
\newblock In {\em H. Greenspan, A. Madabhushi, P. Mousavi, et al. (Eds.), Proceedings of the 26th International Conference on Medical Image Computing and Computer-Assisted Intervention}, pages 35--45. Cham:Springer.

\bibitem[\protect\astroncite{Silva et~al.}{2014}]{silva2014toward}
Silva, J., Histace, A., Romain, O., Dray, X., and Granado, B. (2014).
\newblock Toward embedded detection of polyps in wce images for early diagnosis of colorectal cancer.
\newblock {\em International journal of computer assisted radiology and surgery}, 9:283--293.
\newblock Cham:Springer.

\bibitem[\protect\astroncite{Jha et~al.}{2020}]{jha2020kvasir}
Jha, D., Smedsrud, P., Riegler, M., Halvorsen, P., de~Lange, T., Johansen, D., et~al. (2020).
\newblock {Kvasir-SEG}: A segmented polyp dataset.
\newblock In {\em Y. M. Ro, W.-H. Cheng, J. Kim, et al. (Eds.), Proceedings of the 26th international conference on multimedia modeling}, pages 451--462. Cham:Springer.

\bibitem[\protect\astroncite{S{\'a}nchez-Peralta et~al.}{2020}]{sanchez2020piccolo}
S{\'a}nchez-Peralta, L., Pagador, J., Pic{\'o}n, A., Calder{\'o}n, {\'A}., Polo, F., Andraka, N., et~al. (2020).
\newblock Piccolo white-light and narrow-band imaging colonoscopic dataset: A performance comparative of models and datasets.
\newblock {\em Applied Sciences}, 10(23):8501.
\newblock MDPI.

\bibitem[\protect\astroncite{Ali et~al.}{2023}]{ali2023multi}
Ali, S., Jha, D., Ghatwary, N., Realdon, S., Cannizzaro, R., Salem, O., et~al. (2023).
\newblock A multi-centre polyp detection and segmentation dataset for generalisability assessment.
\newblock {\em Scientific Data}, 10(1):75.
\newblock Nature Publishing Group UK London.

\bibitem[\protect\astroncite{Center}{}]{giana}
Center, R. U.~M.
\newblock Gastrointestinal image analysis ({GIANA}) challenge.
\newblock Retrieved November 10, 2024.
\newblock from: \url{https://endovissub2017-giana.grand-challenge.org/}.

\bibitem[\protect\astroncite{Borji et~al.}{2015}]{borji2015salient}
Borji, A., Cheng, M., Jiang, H., and Li, J. (2015).
\newblock Salient object detection: A benchmark.
\newblock {\em IEEE Transactions on Image Processing}, 24(12):5706--5722.
\newblock Piscataway: IEEE.

\bibitem[\protect\astroncite{Perazzi et~al.}{2012}]{perazzi2012saliency}
Perazzi, F., Kr{\"a}henb{\"u}hl, P., Pritch, Y., and Hornung, A. (2012).
\newblock Saliency filters: Contrast based filtering for salient region detection.
\newblock In {\em Proceedings of the IEEE conference on computer vision and pattern recognition}, pages 733--740. Piscataway: IEEE.

\bibitem[\protect\astroncite{Fan et~al.}{2017}]{fan2017structure}
Fan, D., Cheng, M., Liu, Y., Li, T., and Borji, A. (2017).
\newblock Structure-measure: A new way to evaluate foreground maps.
\newblock In {\em Proceedings of the IEEE international conference on computer vision}, pages 4548--4557. Piscataway: IEEE.

\bibitem[\protect\astroncite{Fan et~al.}{2018}]{Fan2018Enhanced}
Fan, D., Gong, C., Cao, Y., Ren, B., Cheng, M., and Borji, A. (2018).
\newblock Enhanced-alignment measure for binary foreground map evaluation.
\newblock In {\em Proceedings of the 27th International Joint Conference on Artificial Intelligence}, pages 698--704. Cham:Springer.

\bibitem[\protect\astroncite{Zhou et~al.}{2023}]{zhou2023edge}
Zhou, T., Zhang, Y., Chen, G., Zhou, Y., Wu, Y., and Fan, D. (2023).
\newblock Edge-aware feature aggregation network for polyp segmentation.
\newblock {\em arXiv preprint. arXiv:2309.10523}.

\bibitem[\protect\astroncite{Zhao et~al.}{2023}]{zhao2023m}
Zhao, X., Jia, H., Pang, Y., Lv, L., Tian, F., Zhang, L., et~al. (2023).
\newblock {M$^{2}$SNet}: Multi-scale in multi-scale subtraction network for medical image segmentation.
\newblock {\em arXiv preprint. arXiv:2303.10894}.

\bibitem[\protect\astroncite{Zhou et~al.}{2023}]{zhou2023can}
Zhou, T., Zhang, Y., Zhou, Y., Wu, Y., and Gong, C. (2023).
\newblock Can {SAM} segment polyps?
\newblock {\em arXiv preprint. arXiv:2304.07583}.

\bibitem[\protect\astroncite{Shan et~al.}{2021}]{shan2021decouple}
Shan, L., Li, X., and Wang, W. (2021).
\newblock Decouple the high-frequency and low-frequency information of images for semantic segmentation.
\newblock In {\em Proceedings of the IEEE International Conference on Acoustics, Speech and Signal Processing}, pages 1805--1809. Piscataway: IEEE.

\bibitem[\protect\astroncite{Cong et~al.}{2023}]{cong2023frequency}
Cong, R., Sun, M., Zhang, S., Zhou, X., Zhang, W., and Zhao, Y. (2023).
\newblock Frequency perception network for camouflaged object detection.
\newblock In {\em El-Saddik, A.,Mei, T., Cucchiara, R., et al.(Eds.), Proceedings of the 31st ACM International Conference on Multimedia}, pages 1179--1189. New York: ACM.

\bibitem[\protect\astroncite{Wu et~al.}{2021}]{Wu_2021_ICCV}
Wu, H., Chen, G., Wen, Z., and Qin, J. (2021).
\newblock Collaborative and adversarial learning of focused and dispersive representations for semi-supervised polyp segmentation.
\newblock In {\em Proceedings of the IEEE international conference on computer vision}, pages 3489--3498. Piscataway: IEEE.

\bibitem[\protect\astroncite{Cho et~al.}{2023}]{cho2023anti}
Cho, H., Han, Y., and Kim, W. (2023).
\newblock Anti-adversarial consistency regularization for data augmentation: Applications to robust medical image segmentation.
\newblock In {\em H.Greenspan, A. Madabhushi, P.Mousavi, et al.(Eds.), Proceedings of the 26th International conference on medical image computing and computer-assisted intervention}, pages 555--566. Cham:Springer.

\bibitem[\protect\astroncite{Judge et~al.}{2023}]{judge2023asymmetric}
Judge, T., Bernard, O., Cho~Kim, W., Gomez, A., Chartsias, A., and Jodoin, P. (2023).
\newblock Asymmetric contour uncertainty estimation for medical image segmentation.
\newblock In {\em H. Greenspan, A. Madabhushi, P. Mousavi, et al. (Eds.), Proceedings of the 26th International conference on medical image computing and computer-assisted intervention}, pages 210--220. Cham:Springer.

\bibitem[\protect\astroncite{Wei et~al.}{2023}]{wei2023consistency}
Wei, Q., Yu, L., Li, X., Shao, W., Xie, C., Xing, L., et~al. (2023).
\newblock Consistency-guided meta-learning for bootstrapping semi-supervised medical image segmentation.
\newblock In {\em H. Greenspan, A. Madabhushi, P. Mousavi, et al. (Eds.), Proceedings of the 26th International conference on medical image computing and computer-assisted intervention}, pages 183--193. Cham:Springer.

\bibitem[\protect\astroncite{Yin et~al.}{2023}]{yin2023application}
Yin, Z., Yao, C., Zhang, L., and Qi, S. (2023).
\newblock Application of artificial intelligence in diagnosis and treatment of colorectal cancer: A novel prospect.
\newblock {\em Frontiers in Medicine}, 10:1128084.
\newblock Frontiers Media SA.

\bibitem[\protect\astroncite{Palaniappan et~al.}{2024}]{palaniappan2024global}
Palaniappan, K., Lin, E., and Vogel, S. (2024).
\newblock Global regulatory frameworks for the use of artificial intelligence (ai) in the healthcare services sector.
\newblock {\em Healthcare}, 12(5):1730.

\bibitem[\protect\astroncite{Muehlematter et~al.}{2021}]{muehlematter2021approval}
Muehlematter, U., Daniore, P., and Vokinger, K. (2021).
\newblock Approval of artificial intelligence and machine learning-based medical devices in the {USA} and {Europe} (2015--20): a comparative analysis.
\newblock {\em The Lancet Digital Health}, 3(3):e195--e203.
\newblock Elsevier.

\bibitem[\protect\astroncite{Ali et~al.}{2024}]{ali2024assessing}
Ali, S., Ghatwary, N., Jha, D., Isik-Polat, E., Polat, G., Yang, C., et~al. (2024).
\newblock Assessing generalisability of deep learning-based polyp detection and segmentation methods through a computer vision challenge.
\newblock {\em Scientific Reports}, 14(1):2032.
\newblock Nature Publishing Group UK London.

\bibitem[\protect\astroncite{Feng et~al.}{2020}]{feng2020ssn}
Feng, R., Lei, B., Wang, W., Chen, T., Chen, J., Chen, D., et~al. (2020).
\newblock {SSN}: A stair-shape network for real-time polyp segmentation in colonoscopy images.
\newblock In {\em Proceedings of the IEEE 17th International Symposium on Biomedical Imaging}, pages 225--229. Piscataway: IEEE.

\bibitem[\protect\astroncite{Wichakam et~al.}{2018}]{wichakam2018real}
Wichakam, I., Panboonyuen, T., Udomcharoenchaikit, C., and Vateekul, P. (2018).
\newblock Real-time polyps segmentation for colonoscopy video frames using compressed fully convolutional network.
\newblock In {\em K. Schoeffmann, T. H. Chalidabhongse, C.-W. Ngo, et al. (Eds.), Proceedings of the 24th International conference on multimedia modeling}, pages 393--404. Cham:Springer.

\bibitem[\protect\astroncite{Tomar et~al.}{2022}]{tomar2022transresu}
Tomar, N., Shergill, A., Rieders, B., Bagci, U., and Jha, D. (2022).
\newblock {TransResU-Net}: Transformer based {R}es{U}-{N}et for real-time colonoscopy polyp segmentation.
\newblock {\em arXiv preprint. arXiv:2206.08985}.

\bibitem[\protect\astroncite{Wu et~al.}{2021}]{wu2021precise}
Wu, H., Zhong, J., Wang, W., Wen, Z., and Qin, J. (2021).
\newblock Precise yet efficient semantic calibration and refinement in convnets for real-time polyp segmentation from colonoscopy videos.
\newblock In {\em Proceedings of the AAAI conference on artificial intelligence}, volume 35(4), pages 2916--2924. Palo Alto: AAAI Press.

\bibitem[\protect\astroncite{Jha et~al.}{2021}]{jha2021realtime}
Jha, D., Ali, S., Tomar, N., Johansen, H., Johansen, D., Rittscher, J., et~al. (2021).
\newblock Real-time polyp detection, localization and segmentation in colonoscopy using deep learning.
\newblock {\em IEEE Access}, 9:40496--40510.
\newblock Piscataway: IEEE.

\bibitem[\protect\astroncite{Liu et~al.}{2021}]{liu2021feddg}
Liu, Q., Chen, C., Qin, J., Dou, Q., and Heng, P. (2021).
\newblock {FedDG}: Federated domain generalization on medical image segmentation via episodic learning in continuous frequency space.
\newblock In {\em Proceedings of the IEEE conference on computer vision and pattern recognition}, pages 1013--1023. Piscataway: IEEE.

\bibitem[\protect\astroncite{Panigutti et~al.}{2021}]{panigutti2021fairlens}
Panigutti, C., Perotti, A., Panisson, A., Bajardi, P., and Pedreschi, D. (2021).
\newblock Fairlens: Auditing black-box clinical decision support systems.
\newblock {\em Information Processing \& Management}, 58(5):102657.
\newblock Elsevier.

\bibitem[\protect\astroncite{Thambawita et~al.}{2022}]{thambawita2022singan}
Thambawita, V., Salehi, P., Sheshkal, S., Hicks, S., Hammer, H., Parasa, S., et~al. (2022).
\newblock Singan-seg: Synthetic training data generation for medical image segmentation.
\newblock {\em PloS one}, 17(5):e0267976.
\newblock Public Library of Science San Francisco, CA USA.

\bibitem[\protect\astroncite{Pesapane et~al.}{2018}]{pesapane2018artificial}
Pesapane, F., Volont{\'e}, C., Codari, M., and Sardanelli, F. (2018).
\newblock Artificial intelligence as a medical device in radiology: ethical and regulatory issues in europe and the united states.
\newblock {\em Insights into imaging}, 9:745--753.
\newblock Springer.

\bibitem[\protect\astroncite{Zhang and Zhang}{2023}]{zhang2023ethics}
Zhang, J. and Zhang, Z. (2023).
\newblock Ethics and governance of trustworthy medical artificial intelligence.
\newblock {\em BMC Medical Informatics and Decision Making}, 23(1):7.
\newblock Springer.

\end{thebibliography}

\end{document}